\documentclass{article}

\usepackage[T1]{fontenc}
\usepackage[utf8]{inputenc}   
\usepackage{lmodern}

\usepackage[margin=1in]{geometry}

\usepackage{amsmath,amssymb,amsfonts,amsthm}
\usepackage{mathtools}

\usepackage{graphicx}
\usepackage{float}
\usepackage{subcaption}
\usepackage{caption}
\usepackage{pdfpages}
\usepackage{wrapfig}

\usepackage{booktabs}
\usepackage{array}
\usepackage{multirow}

\usepackage[numbers,sort&compress]{natbib}

\usepackage{xcolor}
\usepackage[
    colorlinks=true,
    linkcolor=blue,
    citecolor=blue,
    urlcolor=blue
]{hyperref}
\usepackage[capitalize,noabbrev]{cleveref}

\usepackage{enumitem}
\usepackage{placeins}   
\usepackage{adjustbox}

\theoremstyle{plain}

\theoremstyle{definition}

\theoremstyle{remark}

\hypersetup{
    pdftitle={},
    pdfauthor={},
    pdfsubject={},
    pdfkeywords={}
}

\title{Reconstructing Template-Memorized Images from Natural Prompts}

\author{
Sol Yarkoni\\
School of Electrical \& Computer Engineering\\
Tel Aviv University\\
\texttt{solyarkoni@mail.tau.ac.il}
\and
Mahmood Sharif\\
School of Computer Science \& AI\\
Tel Aviv University\\
\texttt{mahmoods@tauex.tau.ac.il}
\and
Roi Livni\\
School of Electrical \& Computer Engineering\\
Tel Aviv University\\
\texttt{rlivni@tauex.tau.ac.il}
}

\begin{document}

\maketitle
\begin{abstract}
Recent advances in generative models, such as diffusion models, have raised several risks and concerns related to privacy, copyright infringement, and data stewardship. To better understand and mitigate these risks, prior work has proposed techniques and attacks that reconstruct images, or parts of images, from the training set. While these approaches demonstrate that training data can be recovered, they often rely on substantial computational resources, access to the training set, or carefully engineered prompts.

In this work, we devise a new attack that requires low resources, assumes little to no access to the training data, and identifies seemingly benign prompts that lead to potentially risky image reconstruction. We further show that such reconstructions may occur unintentionally and can be produced by users without specific expertise. For example, we observe that, for one existing model, the prompt ``blue Unisex T-Shirt'' generates the face of a real individual. Moreover, by combining the identified vulnerabilities with real-world prompt data, we uncover prompts that reproduce memorized elements.

Our method builds on intuitions from prior work and leverages domain knowledge to reveal a fundamental vulnerability arising from the use of scraped data from e-commerce platforms, where templated layouts and images are associated with pattern-like prompts.
\end{abstract}

\section{Introduction}
\label{sec:intro}

\begin{figure}[t]
\centering
\setlength{\tabcolsep}{3pt}  
\renewcommand{\arraystretch}{1.0}

\begin{tabular}{cc}
\textbf{Source} & \textbf{Generated} \\
\includegraphics[width=0.43\columnwidth]{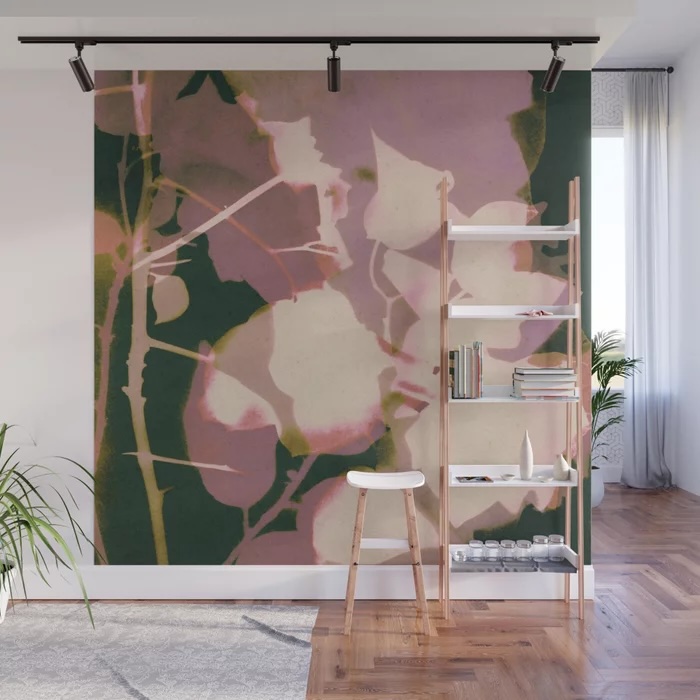} &
\includegraphics[width=0.43\columnwidth]{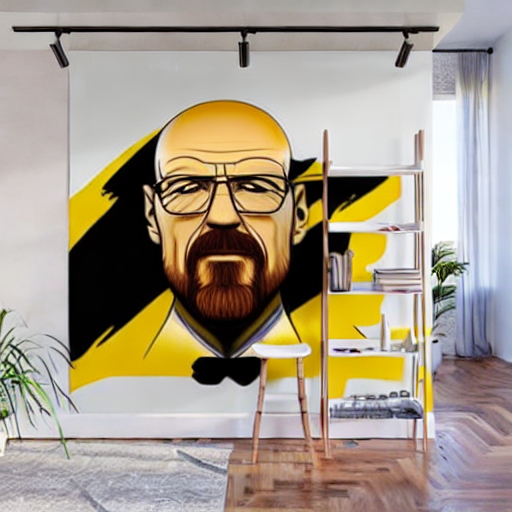} \\
\end{tabular}

\caption{
An image that we generated on a SD 1.4 with the prompt "Walter White wall mural" that was extracted from the database of real-world used prompts DiffusionDB \cite{SDprompts} (right). On its left a corresponding source image the contains elements that were copied. See \cref{ssec:prompts_in_the_wild} for further details.
}
\label{fig:wild_prompt_example}
\end{figure}

With the rising popularity of generative and foundation models, large amounts of public and sometimes private data are being used for training. This widespread use of data raises pressing concerns about privacy and copyright, as models may inadvertently reproduce protected content or reveal sensitive information. As such models become increasingly embedded in everyday applications, addressing these risks has become an urgent challenge for the field. In addition to concerns regarding copyright, the use of public data also raises important questions about consent and context \cite{tramer2024position}. Individuals who have consented to use their photos in a specific setting may not anticipate, nor approve of, their data being repurposed in unrelated or inappropriate contexts.

In turn, researchers are now investigating how data is memorized within generative and foundational models, and whether data can be extracted in ways unintended. Several works \cite{somepalli2023b, carlini2023, webster2023} demonstrated that \emph{attacks} can be designed to extract, \emph{blatantly and verbatim}, data that appeared in the training sets. 

These existing attacks typically rely on access to the training data, substantial computational resources, and the use of specific prompts drawn from that data to trigger extraction. In that sense, they simulate a malicious adversary explicitly attempting to recover training content. However, a critical concern lies in the potential for \emph{unintentional image reconstruction}, where a user issues a \emph{benign} prompt that inadvertently reproduces training data.

\begin{figure*}[t]
\centering
\renewcommand{\arraystretch}{1.1} 
\setlength{\tabcolsep}{6pt}       

\begin{tabular}{>{\centering\arraybackslash}m{0.22\textwidth}
                >{\centering\arraybackslash}m{0.22\textwidth}
                >{\arraybackslash}m{0.46\textwidth}}
\toprule
\textbf{Source} & \textbf{Generated} & \textbf{Prompt} \\
\midrule

\includegraphics[height=2.6cm]{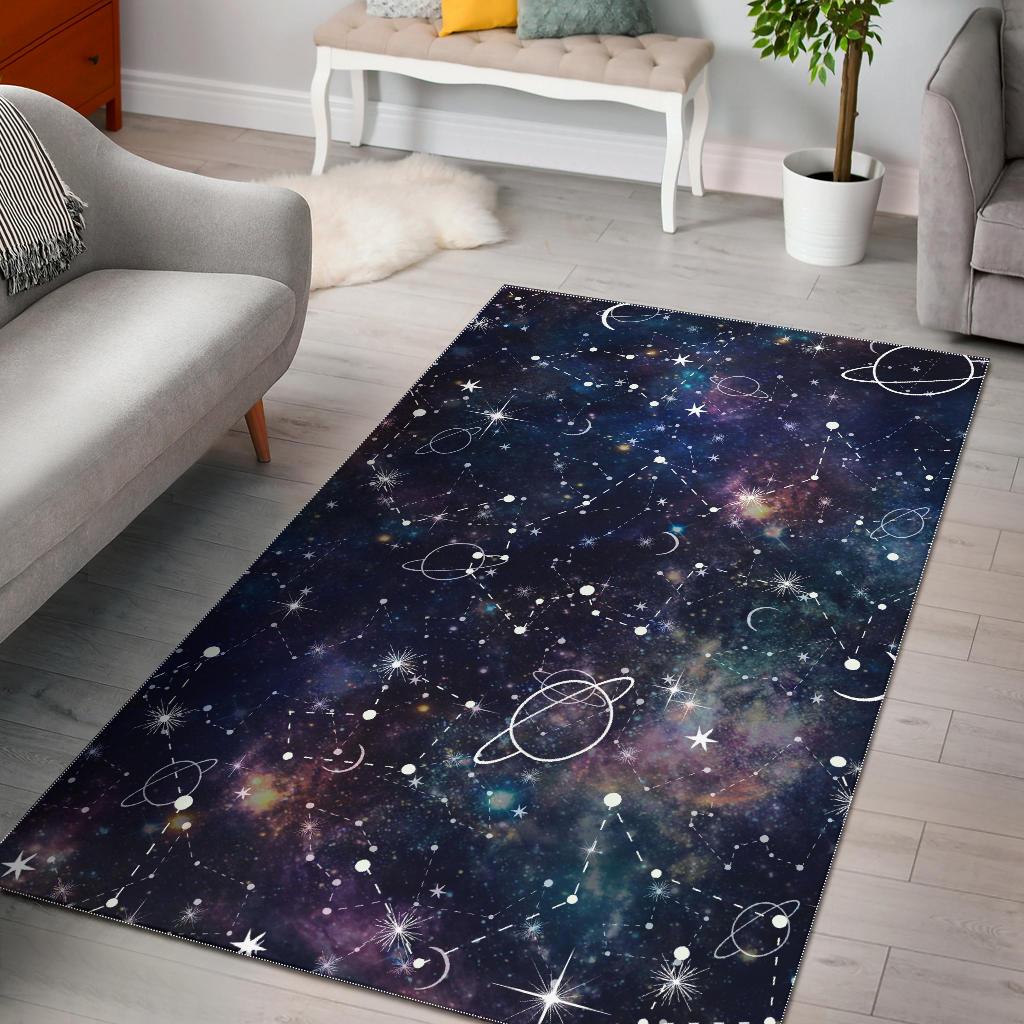} &
\includegraphics[height=2.6cm]{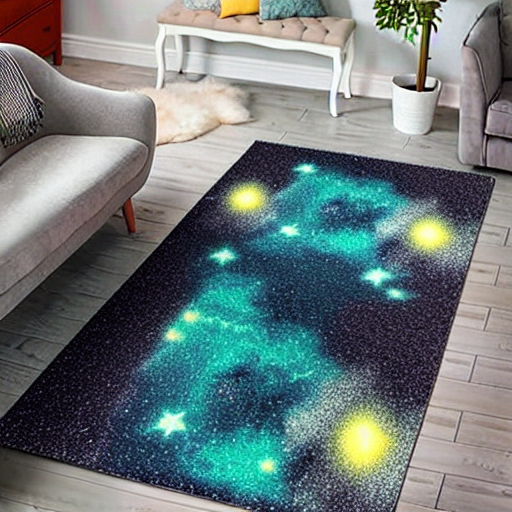} &
"Galaxy Area Rug" \newline (source: GearFrost) \\

\midrule
\includegraphics[height=2.6cm]{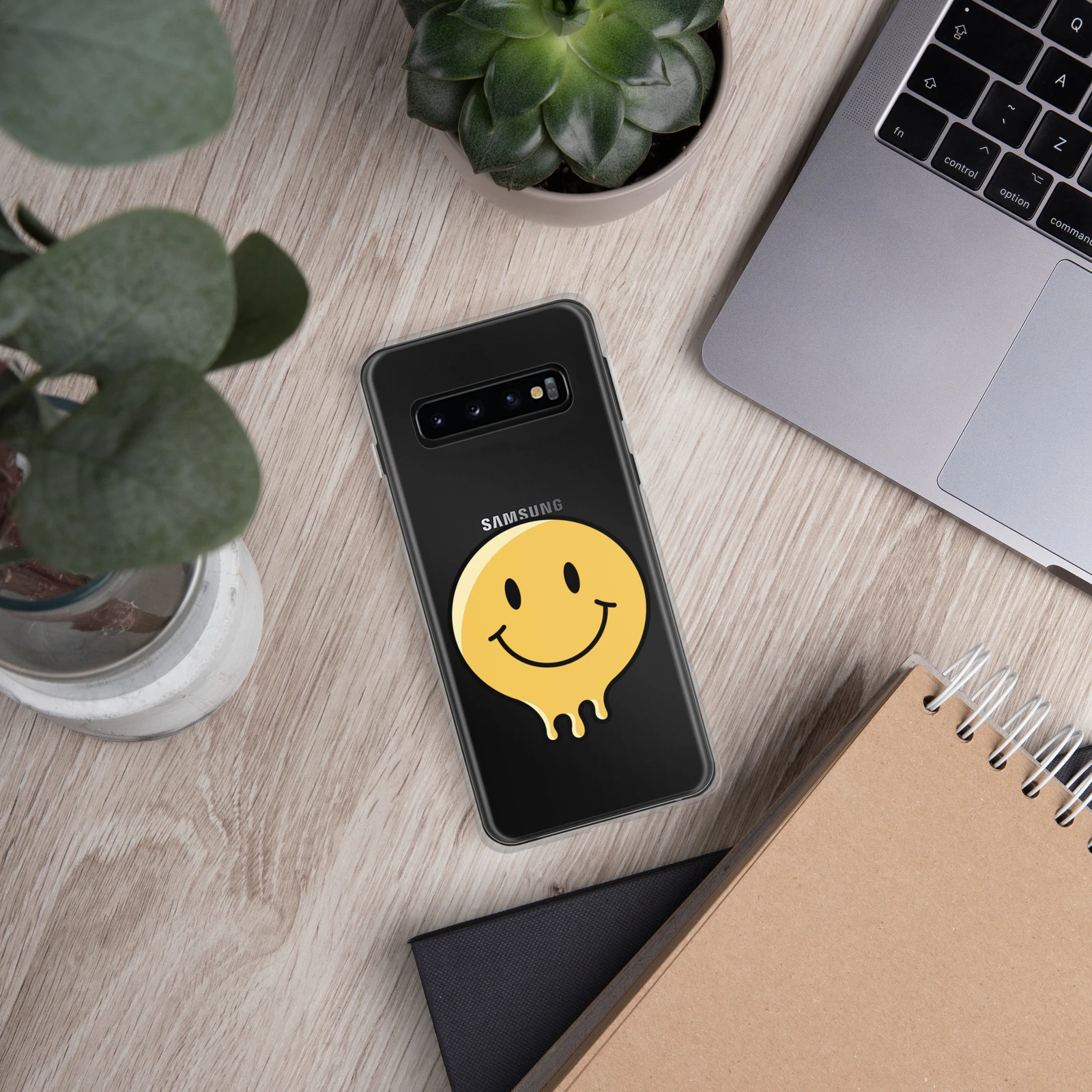} &
\includegraphics[height=2.6cm]{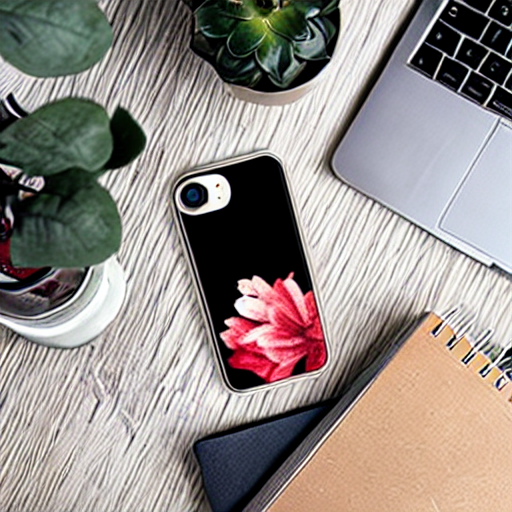} &
"Floral iPhone Case" \newline (source: eBay) \\

\midrule
\includegraphics[height=2.6cm]{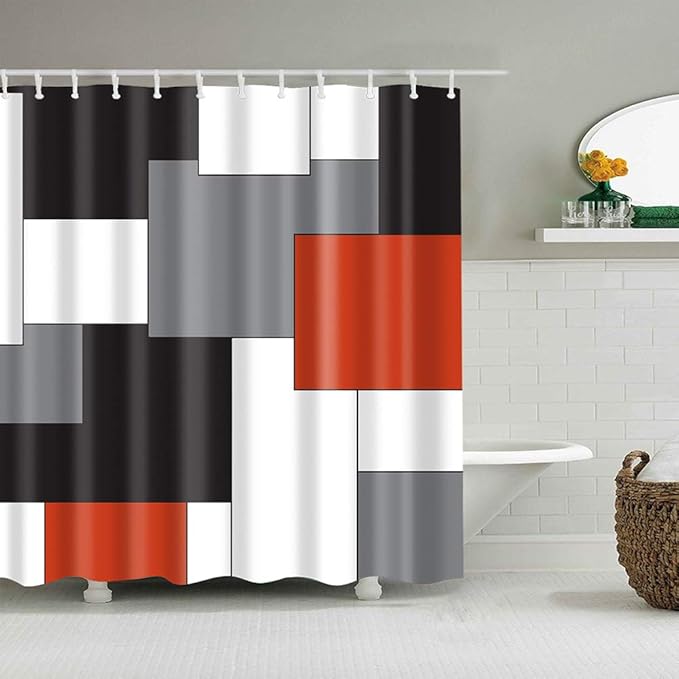} &
\includegraphics[height=2.6cm]{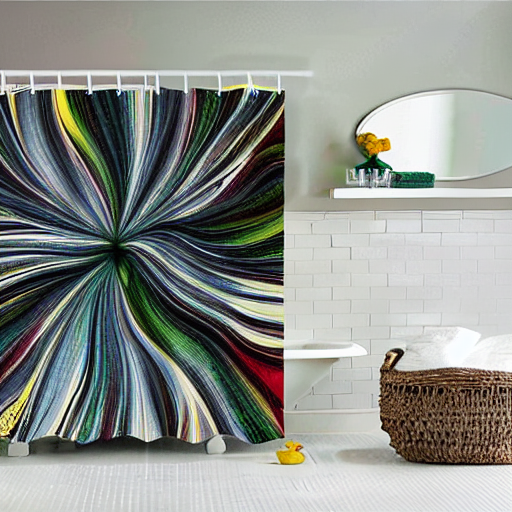} &
"Abstract Art Shower Curtain" \newline (source: Amazon) \\
\bottomrule
\end{tabular}

\vspace{-0.3em}
\caption{
Examples of template-memorized images reconstructed through our attack on SD~1.4.
For each pair, the source image (left), the generated image (center), and the corresponding text prompt (right) are shown.
}
\label{fig:source-generated-prompts}
\end{figure*}

Thus, towards better understanding of the potential risks, we develop in this work a followup attack. Our objective is to construct simple and benign prompts that call for generic objects such as \emph{t-shirt}, \emph{round metal wall art}, \emph{beach towel} etc., without too-specific details. We applied our attack on previously attacked model, and use these prompts to extract images that present elements and objects whose source can be traced to other websites. Perhaps the most disturbing feature of our attack is that real-life models, whose images appeared in such websites, are also extracted by these, so-called innocent prompts, as seen in \cref{tab:reconstruction-examples}. In comparison, previous attacks could extract real images of real people but typically relied on highly specific prompts, typically referring to the individual by name, and the scope was to generate the exact image. For instance, \citet{carlini2023} showed that prompting with “Ann Graham Lotz” could yield a verbatim copy of an existing image. The objective was not to show that her image will be extracted (as the prompt requests that) but to investigate the copying of the existing photo. In contrast, our attack shows that even unintentional prompts can generate images of real individuals without explicitly requesting them (see \cref{tab:reconstruction-examples}). Moreover, using the results from our attack, as well as existing data-bases for real-world used prompts, we could allocate apparently harmless prompts that users indeed entered and can generate images containing copied elements (see \cref{fig:source-generated-prompts}). These behaviors raise distinct concerns around privacy, and individuals' rights. Particularly the right not to have their likeness used for modeling purposes without their consent, or usage of copyrighted material.

Previous work focused on understanding the training data, and in identifying duplications. Our attack avoids harvesting the training data and builds on a working hypothesis that data from \emph{e-commerce} sites and Print on Demand websites (PoDs) is (intentionally or unintentionally) harvested during training. Then, we design an attack that leverages specific traits and domain-knowledge regarding e-commerce sites. 

PoD platforms use automated design placement tools that overlay artwork onto pre-existing images of product by using smart masks and blending techniques. A single platform output may be integrated into many other websites, and the system can instantly generate realistic previews showing the design on different product types, angles, and lighting conditions. As a result, such websites display many images that are identical up to a fixed region where the design is placed. The LAION-5B dataset \cite{schuhmann2022laion5b}, a large-scale, web-scraped image-text dataset, likely contains a significant number of PoD-generated images due to their widespread presence on the internet. We could validate this claim through existing datasets of duplicated images on LAION, \citep{webster2023deduplicationlaion2b}. Importantly though, because generated images are not verbatim copies, we could also validate that many variants of the same image were not necessarily flagged as duplications by these identification systems. In other words, these platforms produce images that may consistently feature repeated elements, such as the same design or product, but with variations in context, angle, background, or lighting. In turn, the images are not always deemed as duplicates, even though they share substantial visual content.

\paragraph{Related Work}
In recent years several reconstruction attacks for foundation models have been designed, including attacks targeted on language models \cite{carlini2021extracting, lee2021deduplicating, kandpal2022deduplicating}  and, more aligned with out work, image reconstruction \cite{somepalli2023b, carlini2023, webster2023}. We elaborate further on these attacks and how they relate to our work in \cref{sec:existing}. 

We are specifically concerned with attacks that generate data from prompts, or \emph{black-box} (to various extent) models that through the ``standard'' use of the model induce types of image conjuring. This is distinct from various attacks that reconstruct training data by analyzing and probing the model \cite{haim2022reconstructing, yin2020dreaming, fredrikson2015model}. These works highlight the memorization of training data by large learning models. Such memorization is necessary for reconstruction attacks to even be possible. There is increasing theoretical support that such memorization may be unavoidable \cite{livni2023information, attias2024information, voitovych2025dichotomy, feldman2020does}, and reconstruction attacks complement these works by showing that such memorization has implications. This leads to the open question of how to mitigate risks and prevent data reconstruction even when the model inherently retains information from its training data.

Given that data is memorized, an important aspect that arises is the question of \emph{originality}, or how to make sure that the output of the model is original and not mere \emph{interpolation} of existing data. This leads to a very interesting theoretical question of what constitute original data \cite{elkin2023can,scheffler2022formalizing}, and to an imminent practical question as to how to identify and regulate non-originality \cite{haviv2024not,hacohen2024not, chiba2025tackling}. Our attack generates results in a middle ground between \emph{blatant, verbatim} copying and \emph{non-copying that lacks originality}, where the system replicates elements without directly copying. While similar interpolations have been observed previously \cite{aithal2024understanding,Somepalli2023}, our method is specifically designed to reliably produce such reconstructions at scale. It is possible that such interpolations can be mitigated and regulated through appropriate credit attribution. The question of how to properly attribute credit, though, is an ongoing research question \cite{livni2024credit}.

Our attack also highlight the risks that can be further amplified through \emph{jail-breaking} methods \cite{ma2024jailbreaking,yang2024sneakyprompt,zhang2024generate}. Jail breaking attacks don't necessarily try to reconstruct data but instead to generate inappropriate usage. We did not attempt to conduct such experiments. However, our attacks generated existing objects and people in new contexts and backgrounds, taken together with such jail-breaking attacks the potential risk should be considered.

\section{Background and Existing Attacks}\label{sec:existing}
The focus of this work is on a new low-resource attack that avoids access to the training set and does not rely on harvesting duplicated data or prompts taken from the training data. While we devise a methodology that can circumvent the use of such access, the intuition behind our attack builds on insights and techniques that were gathered in previous work. Therefore, we briefly review previous works and attacks as well as the important intuition that we build from them.

\paragraph{Random selection}
\Citet{Somepalli2023} demonstrated that by 
analyzing a random subsample of trained data, replicated content can be generated quite straightforwardly. After
designing a detection system for replicated content, they randomly sample 9000 source images from a $12M$ subset of images that were also used in the final rounds of training. Their detection system identified replicated content using the prompts from the training dataset. Interestingly, captions from the dataset don't typically generate the training image, but may conjure a memorized image. 
This attack relies on a small subset on which the model was further trained, and where a typical random sample is duplicated roughly $11$ times \cite{Somepalli2023}. Then the attack generates specific captions that appeared in the dataset verbatim, but they hypothesize that certain key phrases can indeed cause unintentional replicated content. Our work validates this claim and indeed builds on less detailed prompts and short key phrases whose origin were not extracted from the training set, making them much more likely to be used unintentionally.

\paragraph{Training data duplication} 

\Citet{openai2022} collected a dataset of prompts that often resulted in duplicated generated images on their model. They generated images from them, and found the closest in perceptual similarity to an image in the training set. They demonstrated how memorized images had many near-duplicates in the training set. 

\Citet{carlini2023} reasons that training data duplication is a potential cause for  memorization and presents the hypothesis that images extracted from memorization, as opposed to novel generated data, will also contain near-duplicates. 
Their attack takes the most frequently duplicated images along with their captions, generates many images from each caption, and then looks for cliques of nearly identical images. Cliques of at least 10 images are predicted as memorized. 
The clique search uses $\ell_2$ similarity on patches, similar to the split-product method in \cite{Somepalli2023}, and the training data duplicate search relies on entire-image $\ell_2$ and CLIP embedding.
Overall, this attack is resource-intensive and depends on broad access to the training dataset. In addition, the attack is focused on Verbatim copying. Searching for replicated content, we often look for objects in the background or foreground \cite{Somepalli2023}. Thus, an image may be generated only once and still contain replicated content. Identifying such content by inspecting cliques, though, can be challenging.

Another issue is looking for cliques at least 10 images, we've seen empirically that some generated images that we could identify as extracted training images by the similarity to a specific training image, were generated less than 10 times, and even only once. We challenge the implicit assumption of one-to-one memorization, where a specific text is associated with a specific image, causing the specific image to return many times, by showing how some collocations are associated with different image templates, such that the generated images are a mixture of those template along with novel generations.

\paragraph{One-Step Synthesis}
Webster \cite{webster2023} takes a different theoretical approach, selecting the candidates for memorization based on their one-step synthesis behavior, under the hypothesis that memorized image-text pairs present clear, consistent edges after the first denoising step, while non-memorized pairs are blurry after the first denoising step. Using this method they indeed found many captions that extract template memorized images.

To select the candidate captions they relied on full duplicate of image-text pairs. The candidates were selected as the highest scoring 2 million image-text pairs from LAION2B on the duplication metric presented in \cite{webster2023deduplicationlaion2b}. The 2 million candidates include only fully duplicated text-image pairs, and does not include images that were duplicated in a partial fixed region, along with a partial text, but for which no full duplication existed. Out of the 2M candidates, 30k were selected based on their score in the white-box attack presented at \cite{webster2023}, i.e. those for which the most noise had been removed at the first denoising step of Stable Diffusion V1. Due to the method of identifying duplicates, they identified a phenomena of \emph{template memorization}, where only a certain object in the image is being duplicated. 
They are also the first to postulate that such images might be traced to e-commerce sites, a fact that we further validate and helped us in devising our attack that does not require to harvest duplications from the training dataset. Therefore, we expand on the phenomena of template memorization, showing that one could use a natural English text including a short key phrase which we call a \textbf{collocation} rather than a highly specific text, and that such collocations can extract several image templates rather than a specific one. We further rely on e-commerce websites to select candidate prompts and collocations, releasing us from searching and using full duplicates, thus expanding the search to image-text pairs that were duplicated only in part.
\section{Our Attack}\label{sec:our_attack}
\subsection{Data Collection}
Our attack, as in previous work, first builds on extracting potential prompts that will trigger copied works. A distinctive trait of our attack is that we avoid any access to the training set, and want to generate more natural prompts that will lead to reconstructed data. Because we avoid access to the training set, we cannot perform the partial-duplication search described. 
Instead, we collect a list of candidate expressions which are associated with e-commerce websites that were scraped from LAION-5B. We started with an arbitrary list of such sites by asking ChatGPT through the prompt ``consumer facing print on demand websites that appear in LAION 5B.''. Then, we scraped a list of categories and potential sold objects. Other ways to choose the websites ended with similar lists.

For each of these sites, we look at the snapshot of the site from March 2021, since the latest Common Crawl version to enter LAION 5B was April 2021. We then scraped a list of collocations from categories that are sold on these websites such as "unisex t-shirts", ``athletic shoes" etc. Overall we reached a list of around $108$ collocations.
In e-commerce websites, images for each category are generated through different templates. Therefore, in order to generate diversity, that will allow us to identify duplicates, we generated, for each collocation, a set of prompts by adding a description-visual pattern to each. The descriptions were: "Galaxy", "Floral, "Abstract Art" , "I Heart ML", "blue", "red". This process leads to a set of prompts, pattern+collocation phrases, such as "Floral Unisex t-shirts". For each such prompt we generate $25-50$ different images (distinct by their seeds from $0$ to $49$). 

\paragraph{Collections from previous work}
Only for the sake of comparison with previous work, we also extracted 
list of categories that were related to sources that previous work also identified. These categories allow us to further analyze our attack and enable comparisons. 
In more detail, we took the prompts already identified by \citet{webster2023} as what they denote "templates verbatim". These are images memorized up to a fixed mask, or "retrieval verbatim"\footnote{These are identified by generated images in a near duplicate of a training image but the caption of the training image is different than the prompt used to generate the generated image.}. Similarly we extracted collocations from \citet{Somepalli2023}, \citet{hintersdorf24nemo} that we identified as product descriptions (rather than news titles etc.). From these, we manually isolated prompted collocations as the 1-3 words long phrase describing a product. For example, in the list of prompts memorized by Stable Diffusion, published by \citet{webster2023}, appears the prompt "Outshak Hand-Knotted Beige/Blue Area Rug by Shalom Brothers" out of which we extracted the collocation "Area Rug".
After collecting a list of candidate expressions, we generated many images similar to our previous method where we add to the collocation a description pattern.

\subsection{Searching near-duplicates} \label{ssec:clique_search}
Similarly to previous works \cite{openai2022} \cite{carlini2023}, we search for near duplicates among the image generated through the process depicted above. One advantage of our method is that generated images are from known categories. This allows us to use image segmentation pre-trained models to predict the mask of the editable region and filter it out prior to the near-duplicates search. For household items we used \cite{maskFormer}, and for clothing we used \cite{segformerClothesCkpt}.

We then run clique search based on the similarity of the predicted fixed region, where adjacent images are pairs with cosine similarity of above 0.95 on the CLIP embedding.
Clique of at least size 2 were suspected as template-memorized, and were filtered to identify near-duplicates. 
%
In addition, using CLIP instead of $\ell_2$ distance, included images with perturbations which are minor visual variations between images  (see \cref{ssec:perturbations}), that are otherwise very similar in the same clique. 
The method described in this section was clearly limited by the accuracy of the segmentation models, returning only partial results. We leave it for future research to run similar experiments with better segmentation models.
We also searched for near-duplicates through manual visual inspection to complete the results missing from the limitation of the segmentation models.

\begin{figure}[h]
    \setlength{\tabcolsep}{0pt}
    \renewcommand{\arraystretch}{1}
    \begin{tabular}{>{\centering\arraybackslash}m{3.9cm}|| 
    >{\centering\arraybackslash}m{5cm}}
    "X Area Rug" ("rug")&
        \includegraphics[width=.5\textwidth]{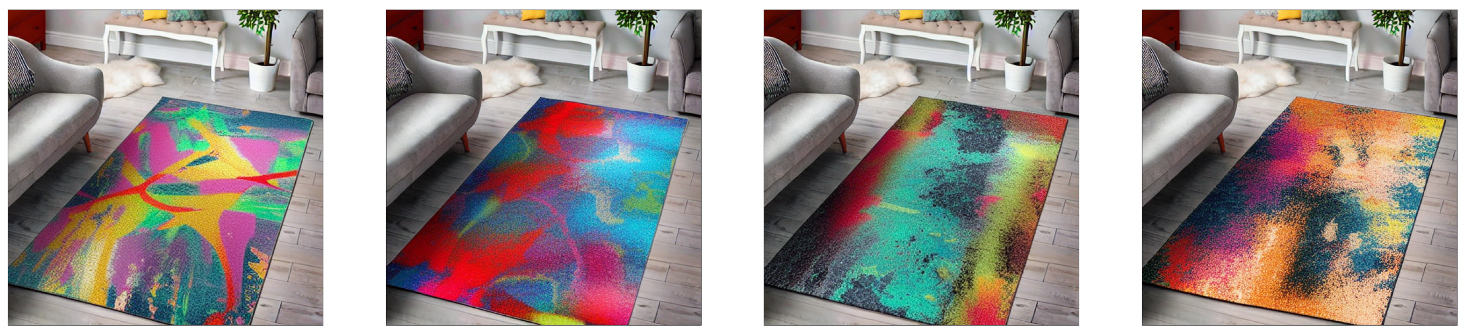} \\
    "X Wall Tapestry" 
("painting")&
        \includegraphics[width=0.5\textwidth]{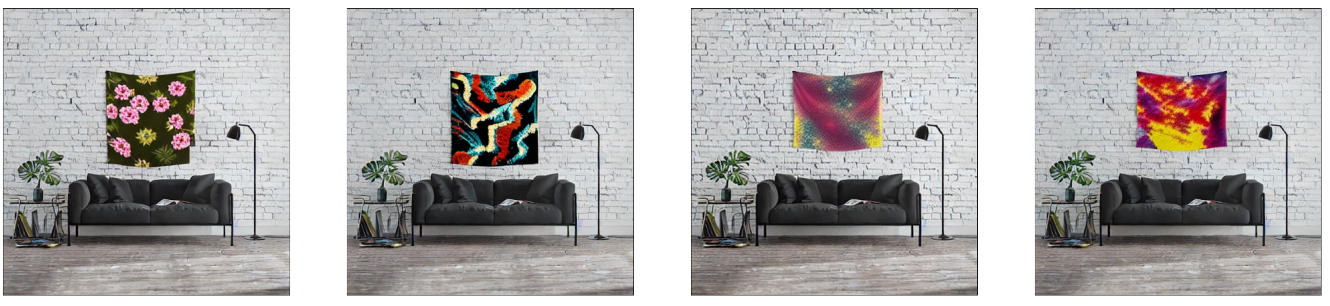} \\
"X Shower Curtain" ("curtain")&        
        \includegraphics[width=0.5\textwidth]{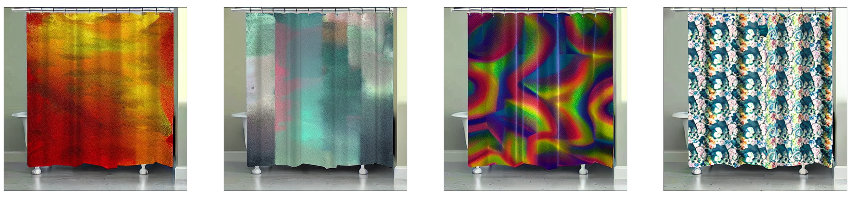} \\
"X High-Tops Sneakers" ("Right-shoe") &
        \includegraphics[width=0.5\textwidth]{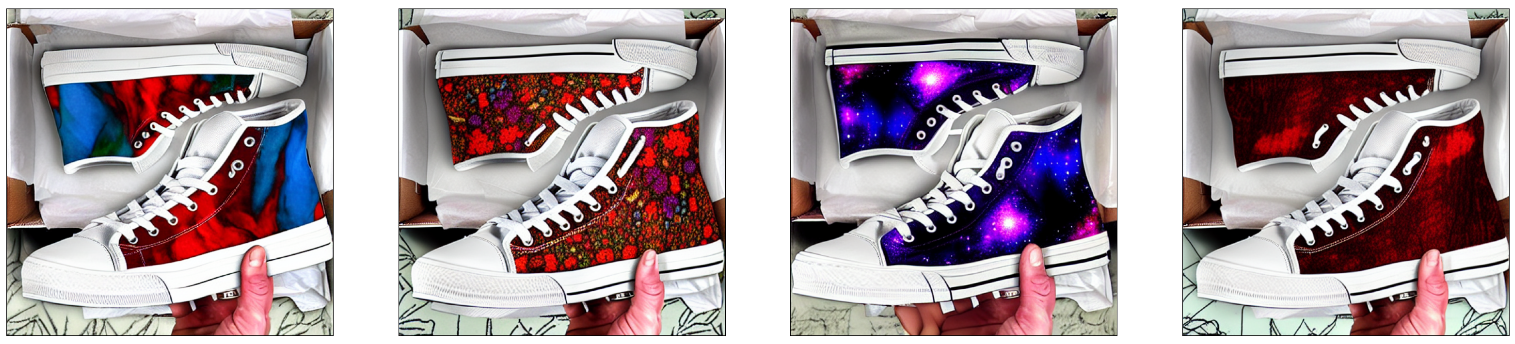} \\
    \end{tabular}
    \caption{
        Examples of image cliques found through our segmentation and masking method. Segmentation category is listed in brackets.
    }
    \label{fig:clique_examples}
\end{figure}

\subsection{Tracing the Source Image}

While for duplicated images we suspect that they are generally copied, for several of the images, we could validate and find a source.
Sources were traced by Google Lens search, by targeting search to e-commerce websites and through visual inspection of images associated with the website categories. 

As we could narrow down the search of the source to handful number of websites, some images were recognized even without duplications, such as the generated image shown in first row of \cref{tab:reconstruction-examples}. For this reason, we then continued to visually inspect the generated images comparing them to the original images in the origin websites, under their product categories, leading to exposing of more such memorized images.
\section{Results}\label{sec:results}

\begin{figure}[h]
    \centering
    \includegraphics[width=0.5\linewidth]{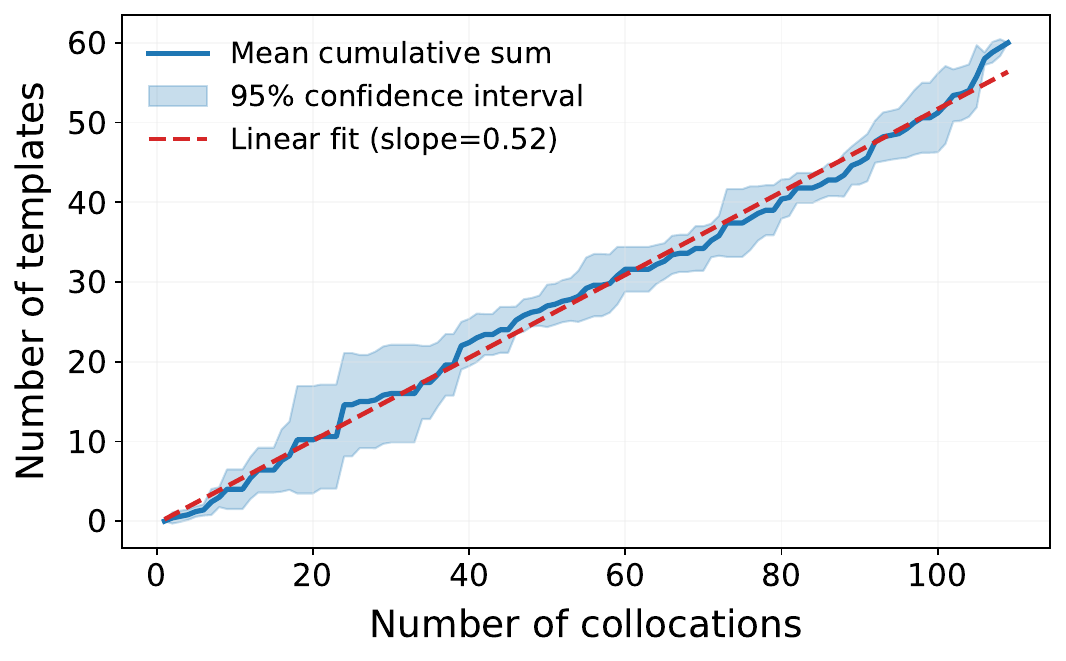}
    \caption{Number of collocations tested vs the number of identified image templates, mean of 5 permutations.}
    \label{fig:collocation_vs_template}
\end{figure}
All of our experiments  were conducted on a single RTX A6000 Machine and, excluding those in \cref{fig:sota_results}, used Stable Diffusion version 1.4 from the official checkpoint accessible at HuggingFace \cite{SD14HuggingFace}, which is the model where previous attacks were introduced \cite{carlini2023,webster2023,somepalli2023b}. Results for more recent models are discussed and reported in \cref{ssec:other_models}.

The 108 collocations that we tested, of which 43 extracted at least one template, produced 67 image templates that we could identify. the dispersion of images is relatively uniform, as is summarized in \cref{fig:collocation_vs_template}. In particular, our method of choosing collocations provides constant marginal returns.
Overall, we generated $11{,}400$ images—comparable to the $9{,}000$ produced by \cite{Somepalli2023}, but orders of magnitude fewer than the $175$ million generated in \cite{carlini2023}. 
Beyond that, whereas prior works focused on individual reconstructions - each counted as a single text–image pair - we identified templates, encompassing a vast range of possible variations enabled by the editability attribute highlighted in this
As shown in \cref{tab:comparison}, compared with previous approaches, our method more accurately pinpoints prompts to reproduce copied images. Finally, we validate our manual procedure for identifying copied images through a user study, summarized in \cref{ssec:user_study}..

\begin{table}[h]
\centering
\small
\renewcommand{\arraystretch}{1.1}
\setlength{\tabcolsep}{2.2pt}
\begin{tabular}{lccc}
\toprule
\textbf{Paper} & \textbf{\# Prompts} & \textbf{\# Variations} & \textbf{\# Copied} \\
\midrule
\citet{carlini2023}     & 350{,}000 & 500 & 109 \\
\citet{Somepalli2023} & 9{,}000   & 1   & 170 \\
\citet{webster2023}     & 30{,}000  & 500 & 153 \\
\textbf{Ours}                           & 108       & 100 & 67  \\
\bottomrule
\end{tabular}
\caption{
Comparison of generated prompts and identified copied images across prior works and ours. “Variations” denotes the number of seeds per prompt in prior works and the average seeds × descriptors per collocation in ours.
}
\label{tab:comparison}
\end{table}

The full list of template groups that we identified, collocations and their associated image template, are provided in listed at \cref{apx:template_groups}. 

\subsection{Images with identified source}
In \cref{fig:clique_examples}, we show examples of four cliques identified across different collocations. Since the cliques are formed using CLIP embeddings rather than raw $\ell_2$ distance, the grouped images may exhibit minor variations such as zoom, color, or pattern changes.

\begin{figure}[t]
\centering
\scriptsize
\renewcommand{\arraystretch}{1.3}
\setlength{\tabcolsep}{1pt} 

\begin{tabular}{
    >{\centering\arraybackslash}m{0.12\columnwidth} ||
    >{\centering\arraybackslash}m{0.14\columnwidth}
    >{\centering\arraybackslash}m{0.14\columnwidth}
    >{\centering\arraybackslash}m{0.14\columnwidth}
    >{\centering\arraybackslash}m{0.14\columnwidth} |
    >{\centering\arraybackslash}m{0.14\columnwidth}
}
\textbf{Category} & \textbf{Abstract} & \textbf{Galaxy} & \textbf{Floral} & \textbf{I Heart ML} & \textbf{Source} \\
\midrule

Beach Towel &
\includegraphics[width=\linewidth]{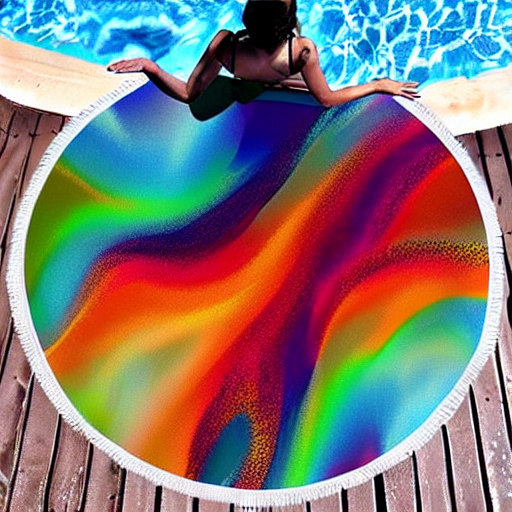} &
\includegraphics[width=\linewidth]{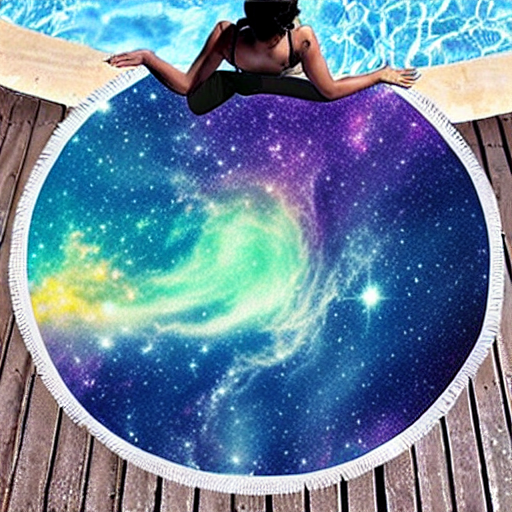} &
\includegraphics[width=\linewidth]{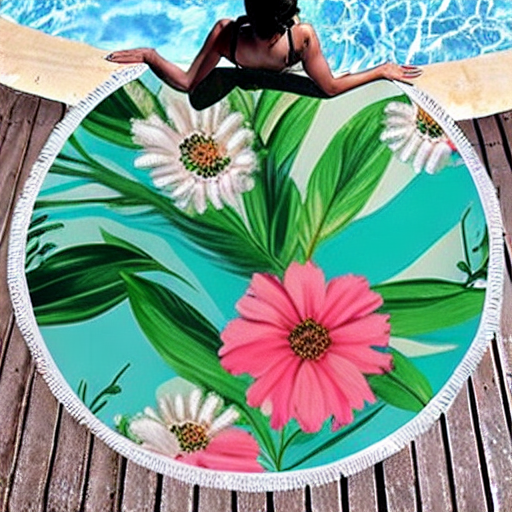} &
\includegraphics[width=\linewidth]{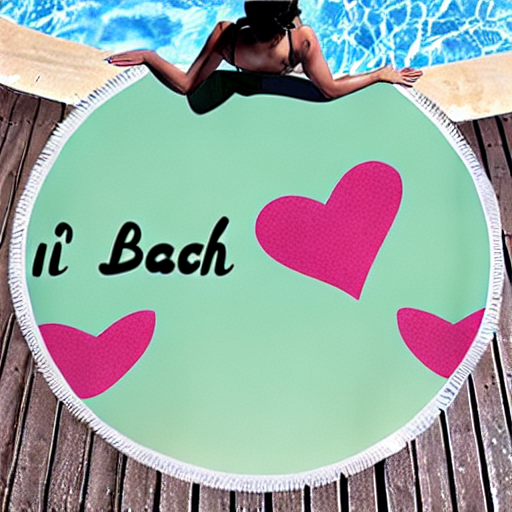} &
\includegraphics[width=\linewidth]{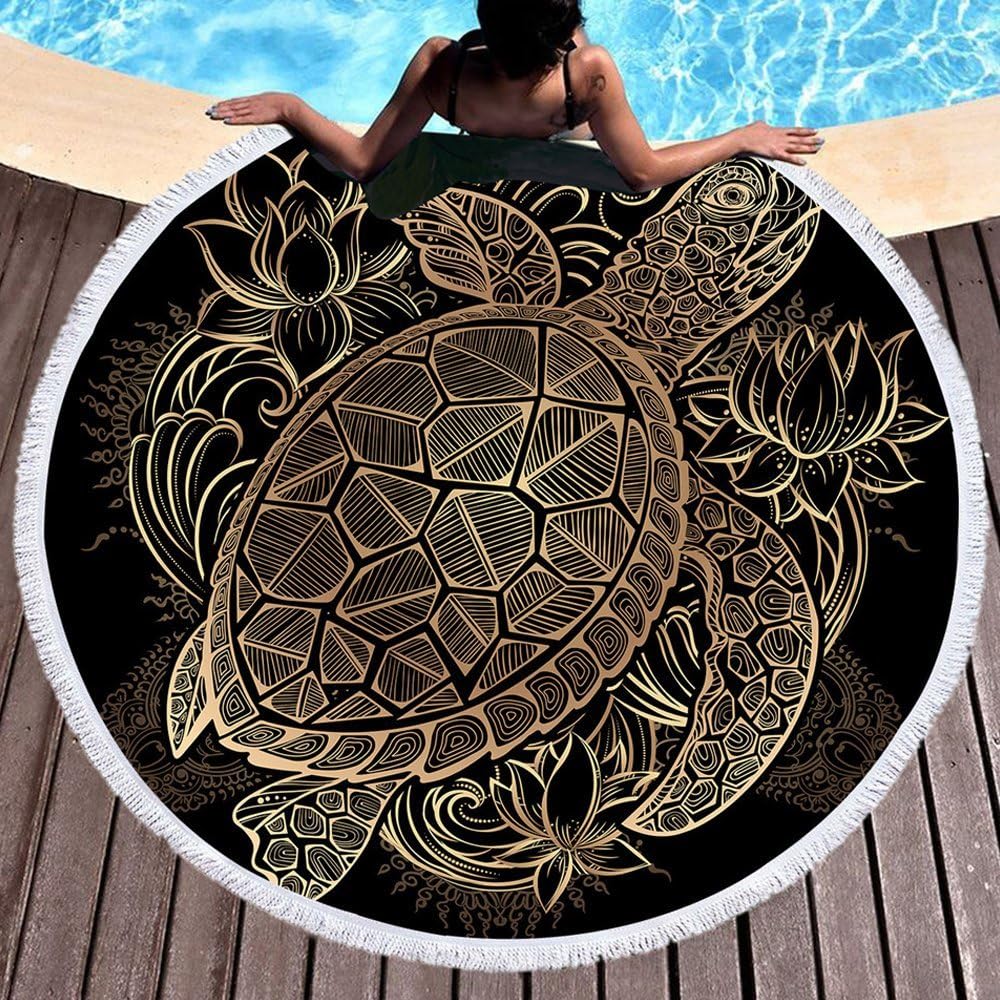} \\

Throw Pillow &
\includegraphics[width=\linewidth]{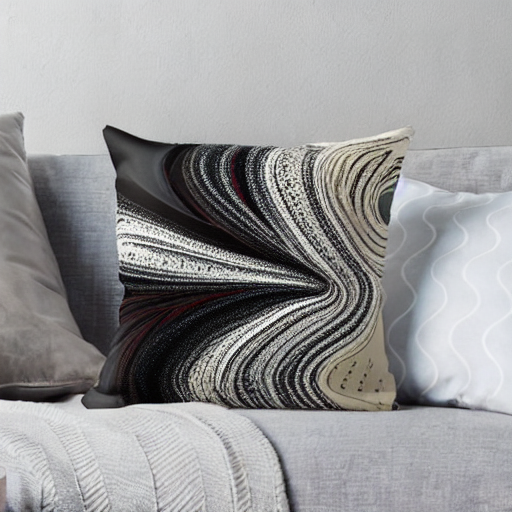} &
\includegraphics[width=\linewidth]{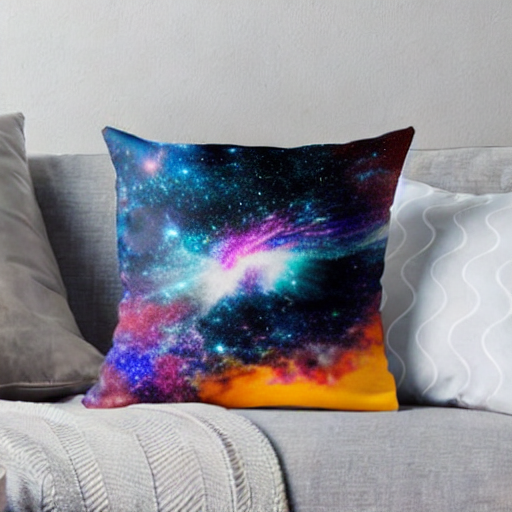} &
\includegraphics[width=\linewidth]{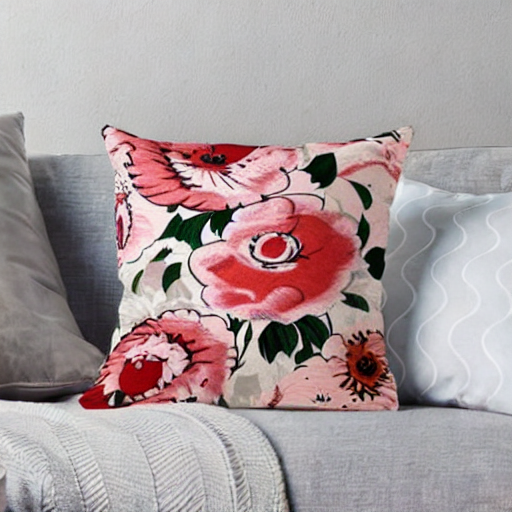} &
\includegraphics[width=\linewidth]{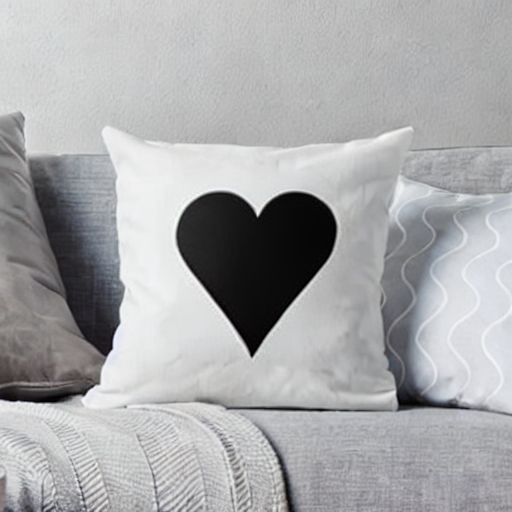} &
\includegraphics[width=\linewidth]{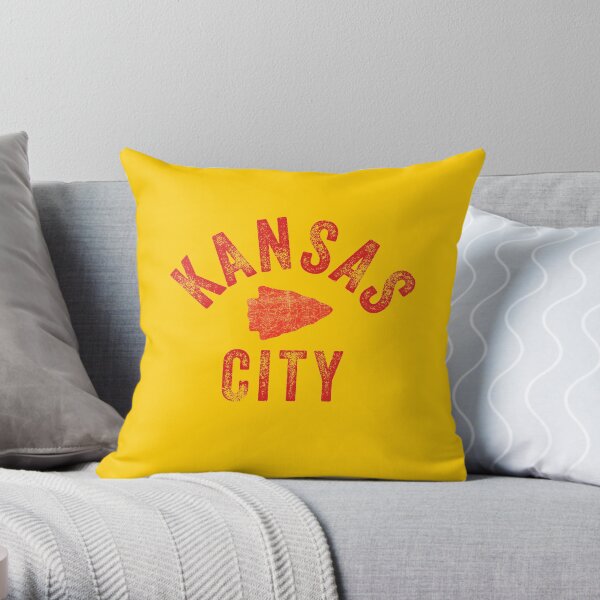} \\

T-Shirt (1) &
\includegraphics[width=\linewidth]{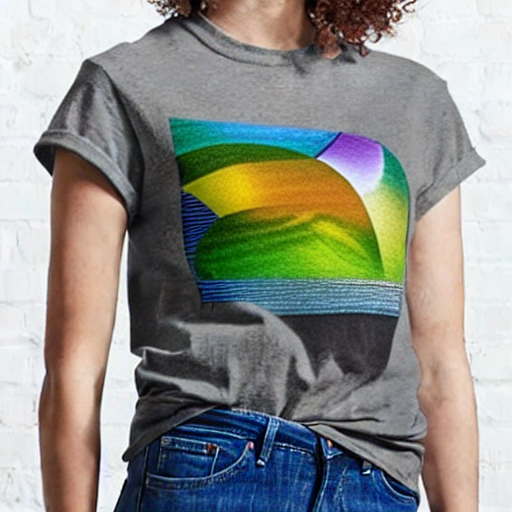} &
\includegraphics[width=\linewidth]{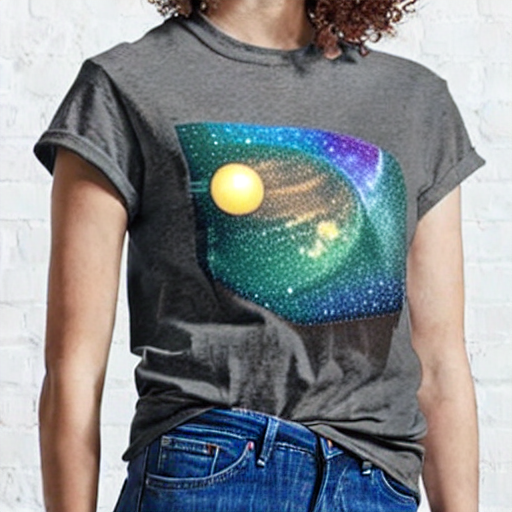} &
\includegraphics[width=\linewidth]{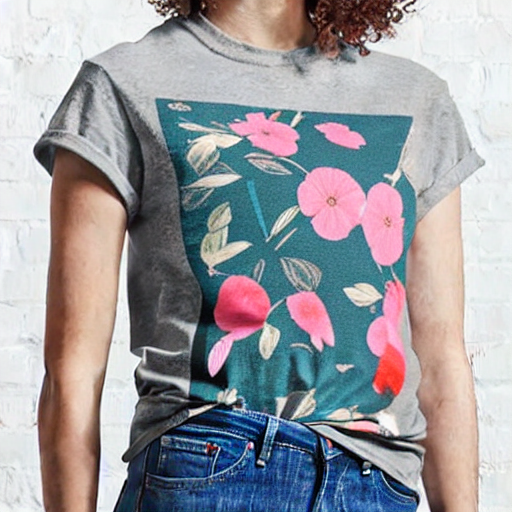} &
\includegraphics[width=\linewidth]{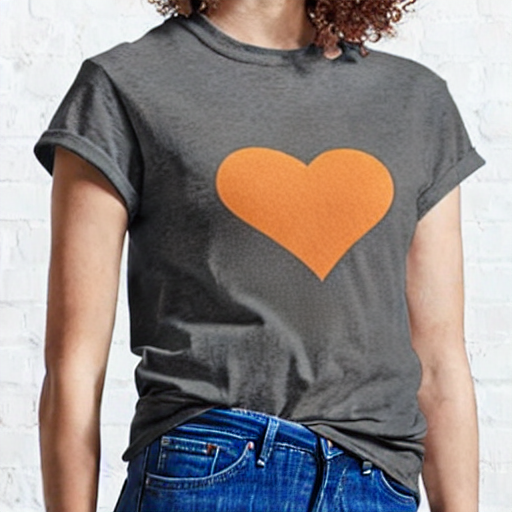} &
\includegraphics[width=\linewidth]{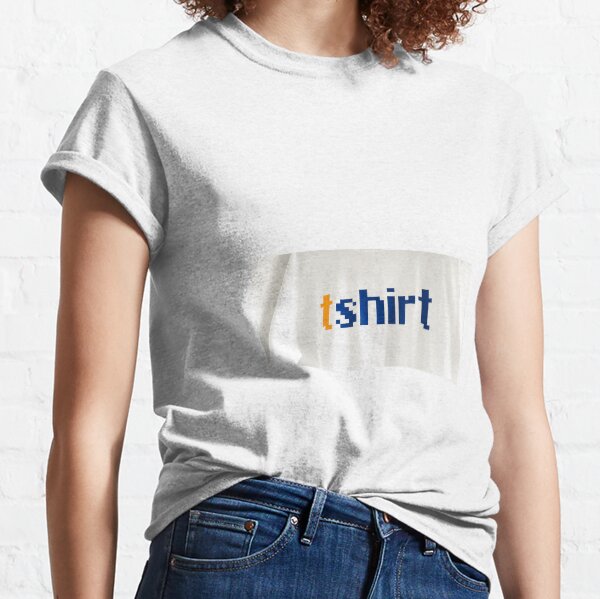} \\

T-Shirt (2) &
\includegraphics[width=\linewidth]{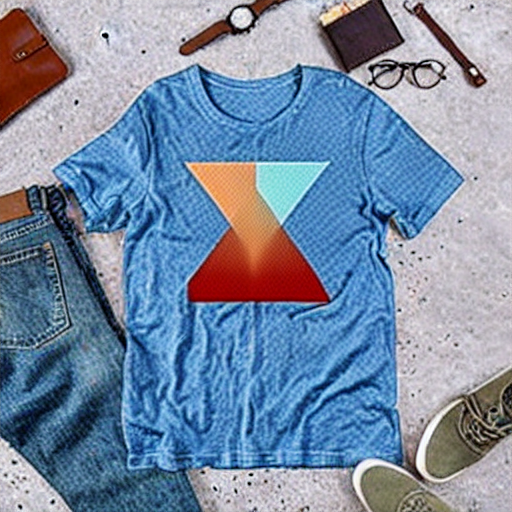} &
\includegraphics[width=\linewidth]{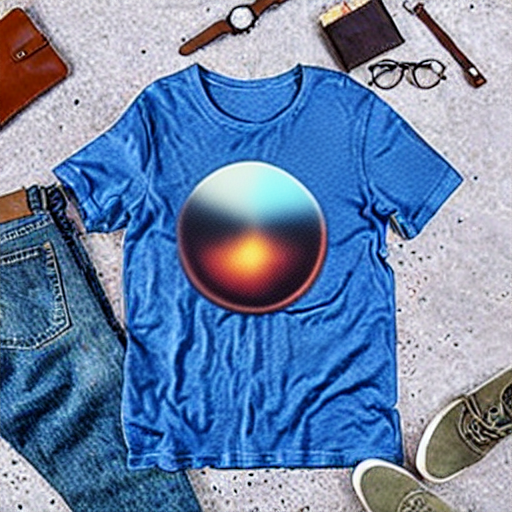} &
\includegraphics[width=\linewidth]{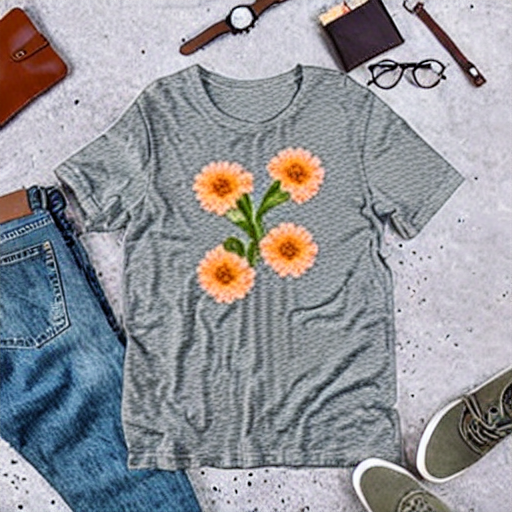} &
\includegraphics[width=\linewidth]{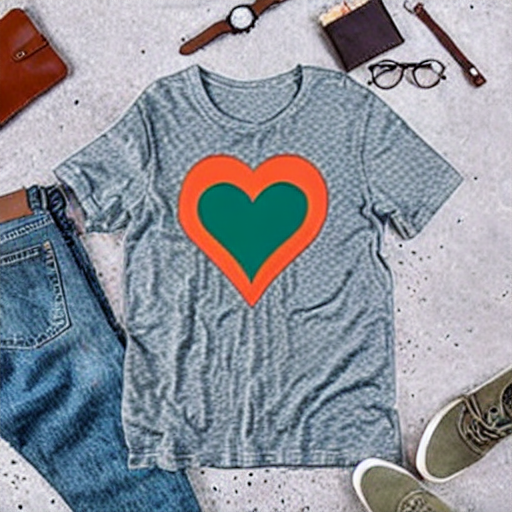} &
\includegraphics[width=\linewidth]{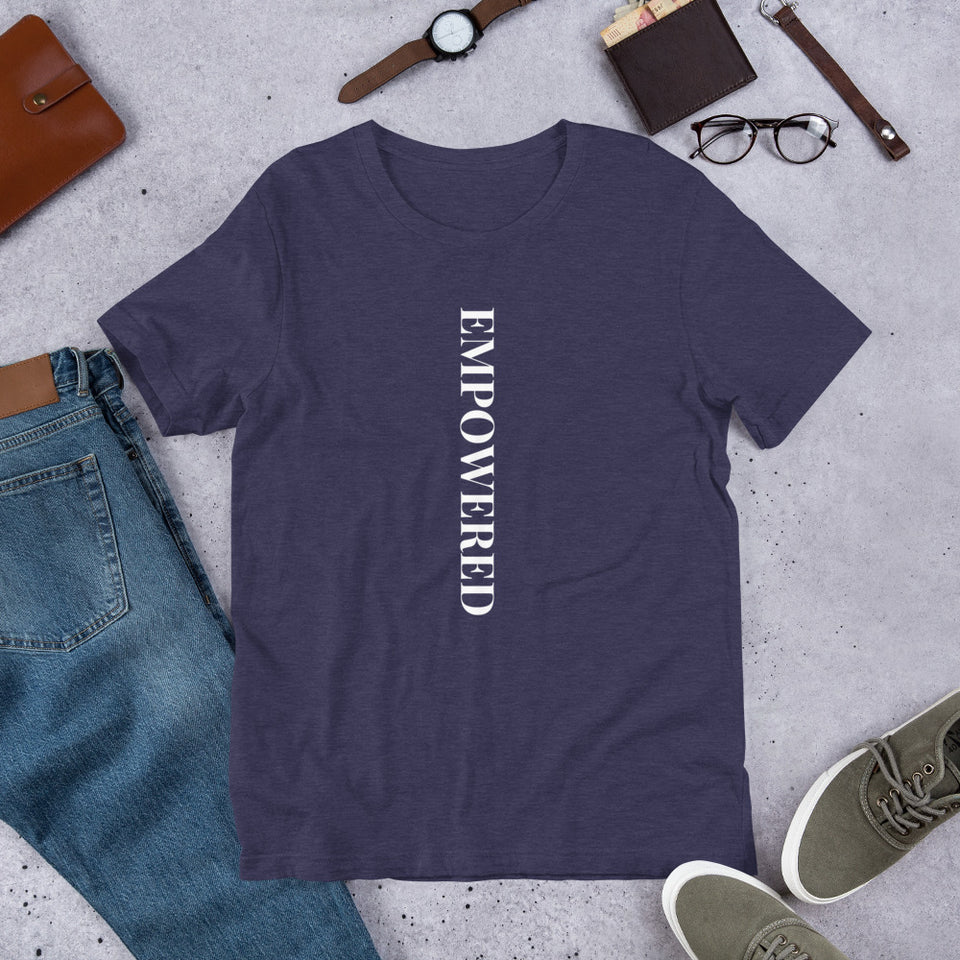} \\

\end{tabular}

\caption{
Examples across product categories, showing different prompt themes (columns) and their corresponding real source image (last column).
}
\label{fig:generic}
\end{figure}

\begin{figure}[H]
\centering
\renewcommand{\arraystretch}{1.5}

\scalebox{0.8}{%
\begin{tabular}{>{\centering\arraybackslash}m{0.5\linewidth} 
                >{\centering\arraybackslash}m{0.38\linewidth}}
\toprule
\textbf{Generated Image \& Source} & \textbf{Prompt Used \& Source} \\
\midrule

\includegraphics[width=.8\linewidth]{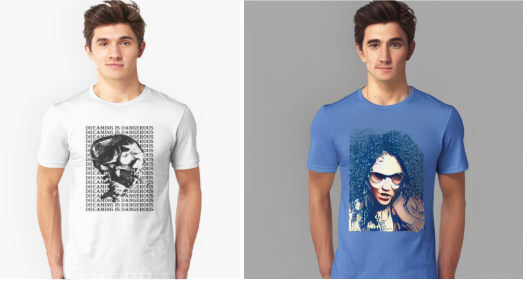}
&
\small{"Abstract Art Unisex T-Shirt" \newline (source: Redbubble)}
\\
\midrule

\includegraphics[width=.8\linewidth]{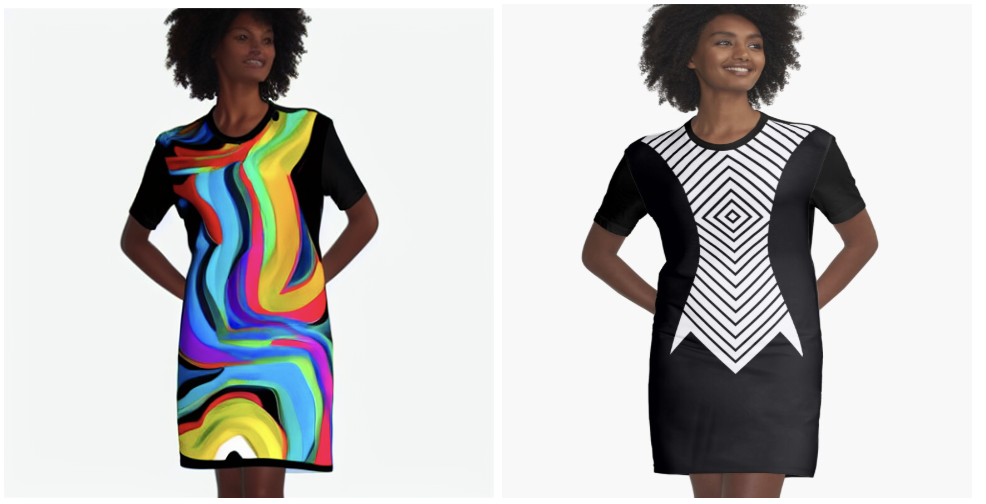}
&
\begin{center}\small{"Abstract Art Graphic T-Shirt Dress" \newline (source: Redbubble)}\end{center}
\\
\midrule

\includegraphics[width=.8\linewidth]{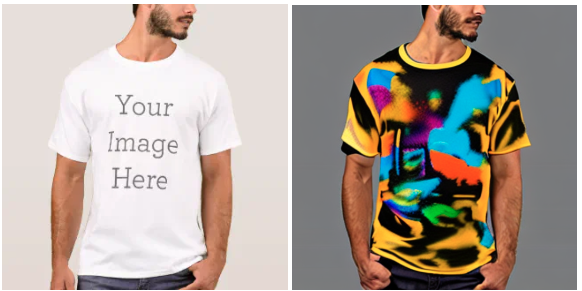}
&
\small{``Abstract Art Essential T-Shirt'' \newline (source: Zazzle)}
\\
\bottomrule
\end{tabular}
}

\caption{
Examples of generated people that we could identify in a source (left: generated image, right: source).
The first row contains a person identified by name \cite{jackstrattinsmith}. Attack conducted on SD~1.4.
}
\label{tab:reconstruction-examples}
\end{figure}

For many duplicated images, we were able to locate their source counterparts on the web, as shown in \Cref{fig:generic}. Interestingly, even for some images not automatically identified as duplicates, we could still trace their origins by inspecting e-commerce sites from which our categories were derived—for example, the image in the first row of \Cref{tab:reconstruction-examples}.

In \Cref{fig:benign_vs_full_prompts}, we show two cases where the source images and their original captions could be identified in LAION, as they were borrowed from prior works with dataset access. The reconstruction prompts we used did not include these captions but rather generic descriptions, underscoring the risk of unintentional image conjuring.

\paragraph{Real Humans}
Arguably, the most concerning type of image recall are those which potentially contain a real human model. Several of the duplicates we identified indeed contain real humans that we could find their source, see \cref{tab:reconstruction-examples}.

As discussed, previous works also managed to extract real-humans \cite{carlini2023}. However this was generally done by prompts that actively request that person's image, and the focus was on copyright and verbatim copy.
Here, the person is extracted by a prompt that, seemingly, does not contain their name, but instead the prompt contains a collocation associated with a product that the model had presented in the image. For example, mundane items such as "T-Shirt". This is both concerning as these models were compensated for generating exactly such images, and as an uninformed user could generate such images under the impression that the human image is synthetic.

Even in cases where the face was deformed (as often happens in SD) or the picture is only of the torso, it might still contain identifying markers such as tattoos as well as other specific visual attributes such as haircut and pose. See  \cref{tab:interpolation} (left columns) for further details.

\subsection{Comparison with previous attacks}\label{sec:compare}
\begin{figure}[]
\centering

\begin{tabular}{>{\centering\arraybackslash}m{0.2\linewidth}
                >{\centering\arraybackslash}m{0.2\linewidth}
                >{\centering\arraybackslash}m{0.4\linewidth}}
\toprule
\textbf{Generated Image} & \textbf{Source Image} & \textbf{ Prompt and Source Caption} \\
\midrule
\includegraphics[height=0.6\linewidth]{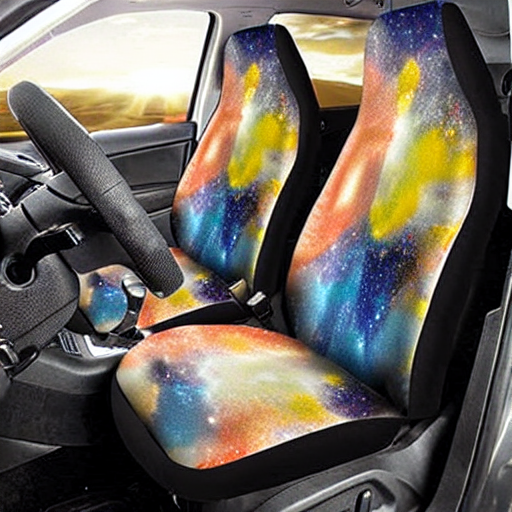} &
\includegraphics[height=0.6\linewidth]{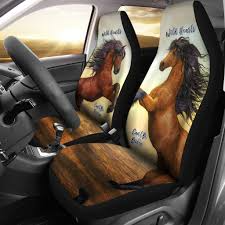} &
\small
\textbf{prompt}: Galaxy Print Universal Fit Car Seat Covers. 
\newline \textbf{Caption}: Wild Hearts Can'T Be Broken Car Seat Covers For Horse Lovers 170804 - YourCarButBetter

\\
\midrule
\includegraphics[height=0.6\linewidth]{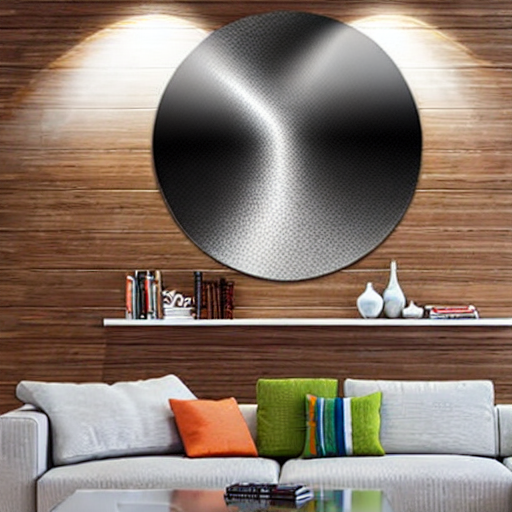} &
\includegraphics[height=0.6\linewidth]{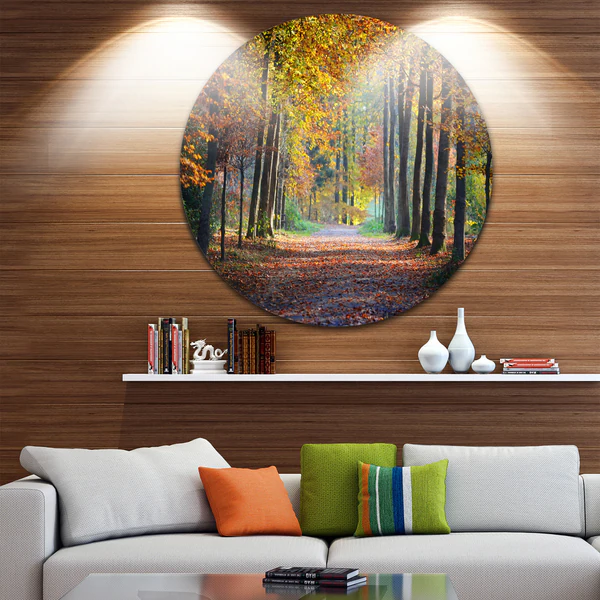} &
\small
\textbf{Prompt}: Abstract Art Round Metal Wall Art.\newline
\textbf{Caption}: Designart 'Wide Pathway in Yellow Fall Forest' Landscape Photo Round Metal Wall Art
. \\

\bottomrule
\end{tabular}
\caption{\small Comparison of Generated and Source Images with Corresponding Prompts and Captions when taking the categories from previous works, examples from \cite{hintersdorf24nemo}.}\label{fig:collocations_from_previous_works}
\end{figure}

As discussed, to allow comparison with previous attacks we also extracted a list of collocations, similar to the one we extracted from the generic websites we chose, by harvesting e-commerce sites that appeared in previous works. For example, we could extract the collocation \emph{car seat cover} and compare our approach to the previous approach that extracts captions from the training set. \cref{fig:collocations_from_previous_works} provide examples of images and the associated prompts used in each attack. 

\begin{figure}[h]
\centering
\footnotesize
\setlength{\tabcolsep}{1.5pt}
\renewcommand{\arraystretch}{1.05}

\newcommand{\cellimg}[1]{%
  \includegraphics[width=\linewidth, height=2.4cm, keepaspectratio]{#1}%
}

\begin{tabular}{
  >{\centering\arraybackslash}m{0.23\columnwidth}
  >{\centering\arraybackslash}m{0.23\columnwidth}
  >{\centering\arraybackslash}m{0.23\columnwidth}
  >{\centering\arraybackslash}m{0.23\columnwidth}
}
\toprule
\multicolumn{2}{c}{\textbf{Webster}} &
\multicolumn{2}{c}{\textbf{Ours}} \\
\cmidrule(lr){1-2} \cmidrule(lr){3-4}
\textbf{1} & \textbf{10} & \textbf{1} & \textbf{10} \\
\midrule

\cellimg{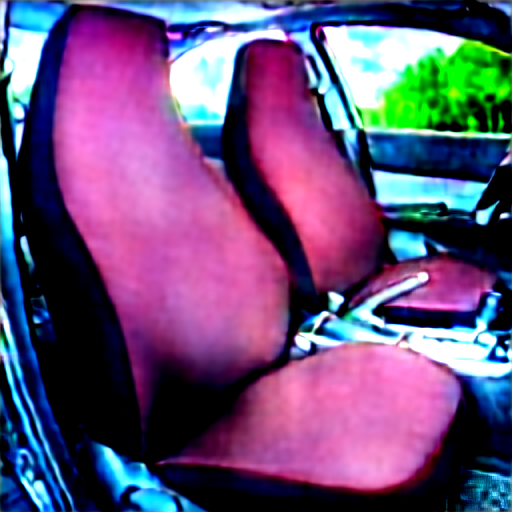} &
\cellimg{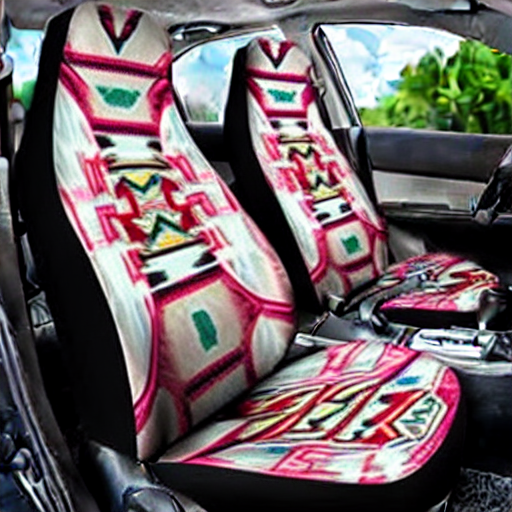} &
\cellimg{example_images/webster-comparison/car_seat_covers_1_step_o.png} &
\cellimg{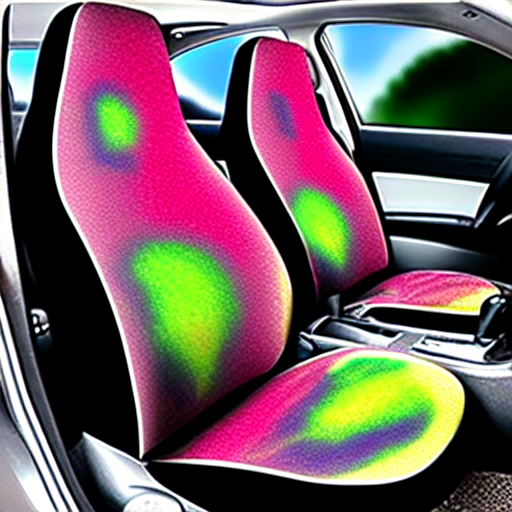} \\
\multicolumn{2}{c}{\emph{Tribal Aztec Indians pattern}} &
\multicolumn{2}{c}{\emph{Floral Car Seat Covers}} \\
\midrule

\cellimg{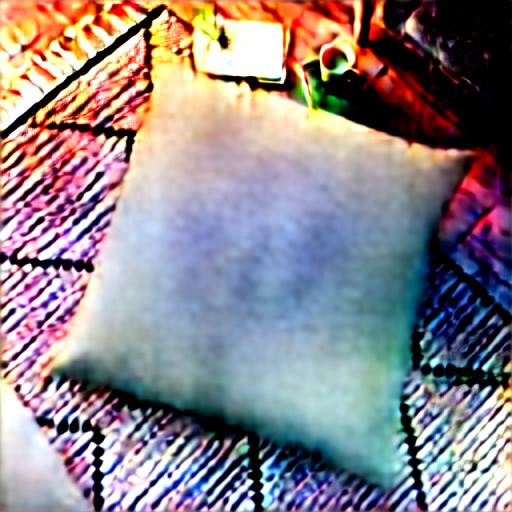} &
\cellimg{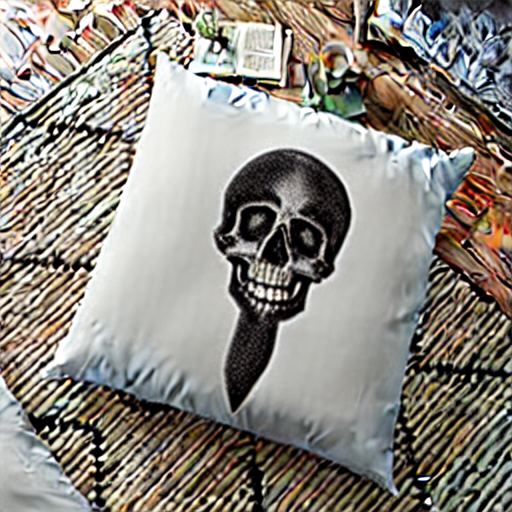} &
\cellimg{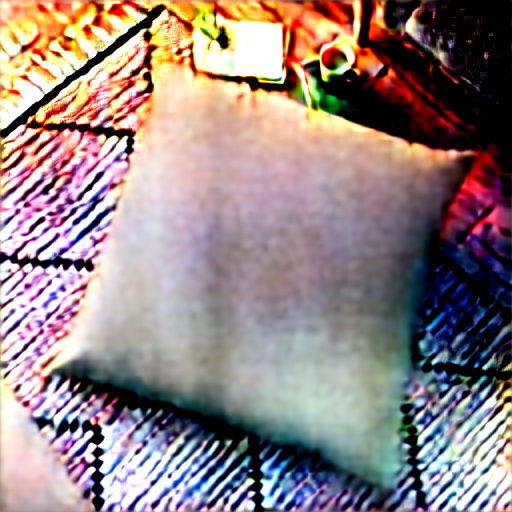} &
\cellimg{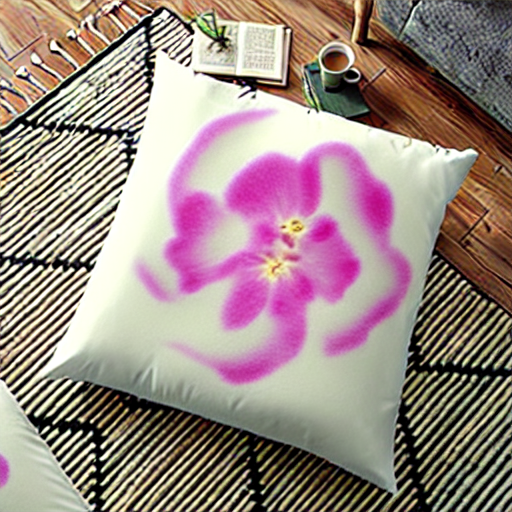} \\
\multicolumn{2}{c}{\emph{Skull Floor Pillow}} &
\multicolumn{2}{c}{\emph{Floral Floor Pillow}} \\
\midrule

\cellimg{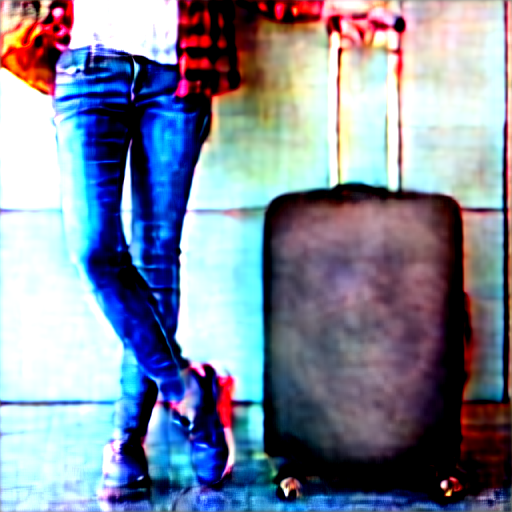} &
\cellimg{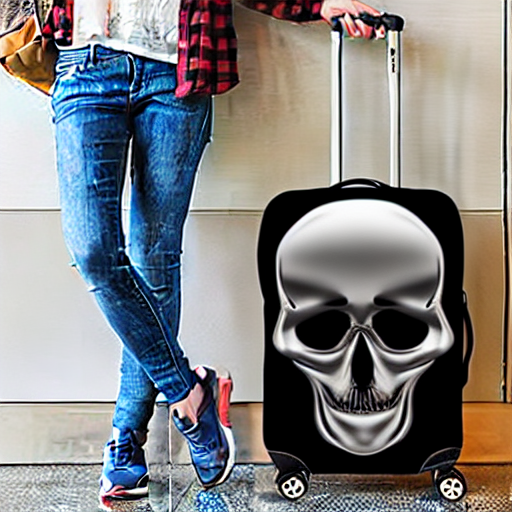} &
\cellimg{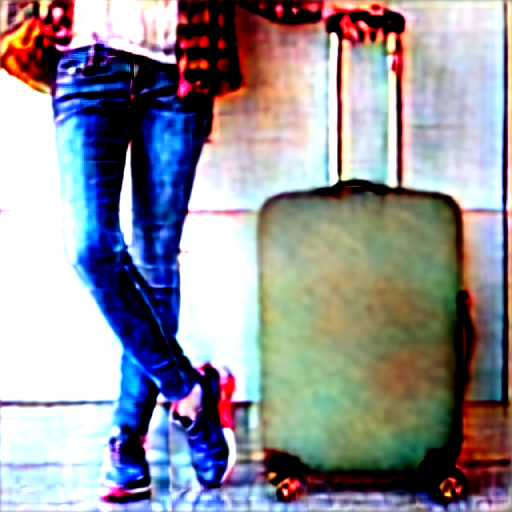} &
\cellimg{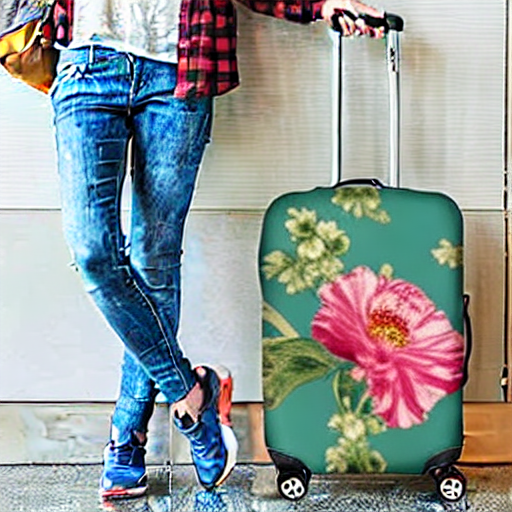} \\
\multicolumn{2}{c}{\emph{3D Skull King Luggage Cover}} &
\multicolumn{2}{c}{\emph{Floral Luggage Cover}} \\
\bottomrule
\end{tabular}

\caption{
Comparison between images generated after 1 and 10 steps for Webster’s attack and ours.
From top to bottom: (a) identified by Webster’s white-box only, (b) identified as non-verbatim, (c) identified as template verbatim.
}
\label{fig:comparison}
\end{figure}

Demonstration of the one-step denoising behavior presented at \cite{webster2023}can be seen at \ref{fig:comparison}.

\subsection{Attacks on State-Of-The-Art and other models} \label{ssec:other_models}

\begin{figure*}[ht]
    \centering
    \setlength{\tabcolsep}{4pt} 

    \begin{tabular}{cccccc}
        \multicolumn{2}{c}{\textbf{Stable Diffusion v3.5}} &
        \multicolumn{2}{c}{\textbf{Flux-Schnell v1.0}} &
        \multicolumn{2}{c}{\textbf{MidJourney v6.1}} \\

        \textbf{Source} & \textbf{Generated} &
        \textbf{Source} & \textbf{Generated} &
        \textbf{Source} & \textbf{Generated} \\

        \includegraphics[width=0.13\linewidth]{example_images/sd35/shower_curtain_wo_duck.jpg} &
        \includegraphics[width=0.13\linewidth]{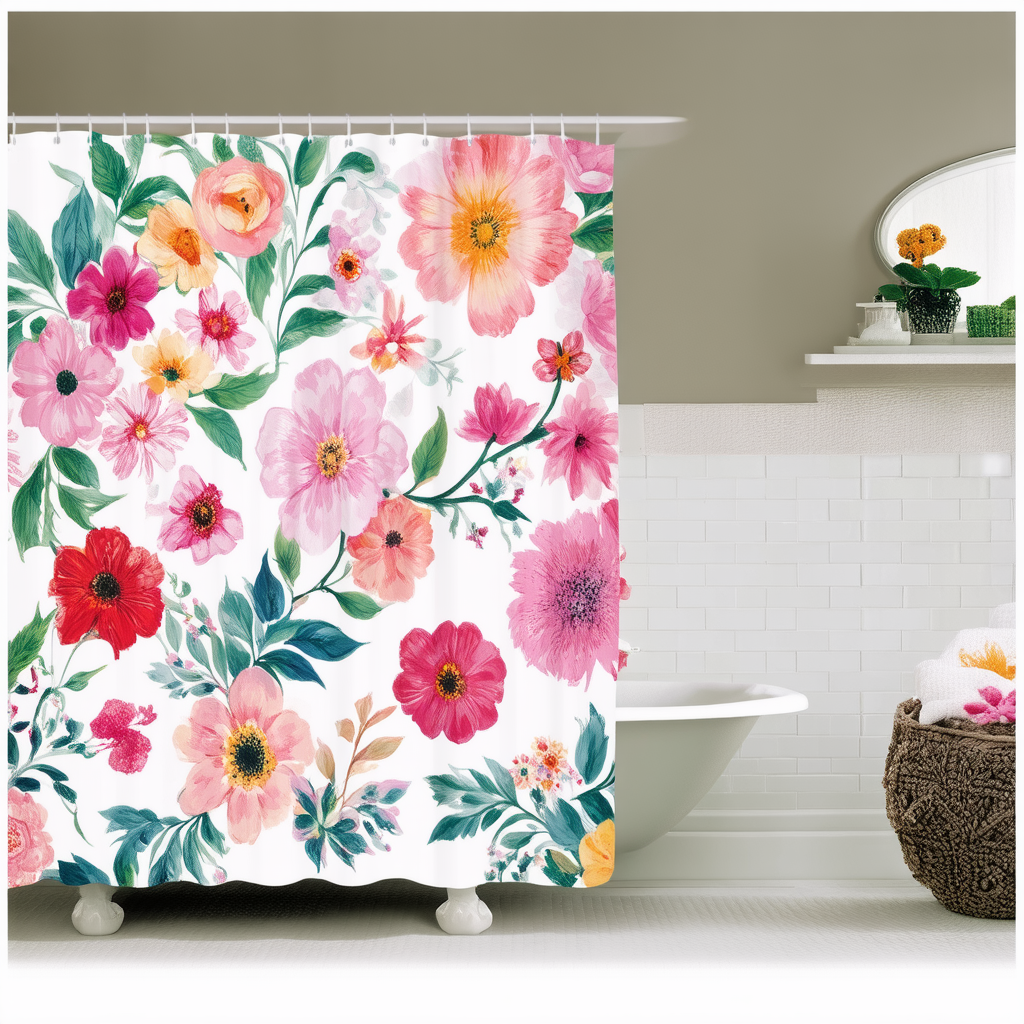} &

        \includegraphics[width=0.13\linewidth]{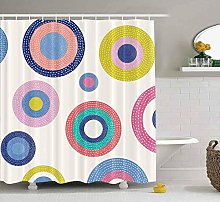} &
        \includegraphics[width=0.13\linewidth]{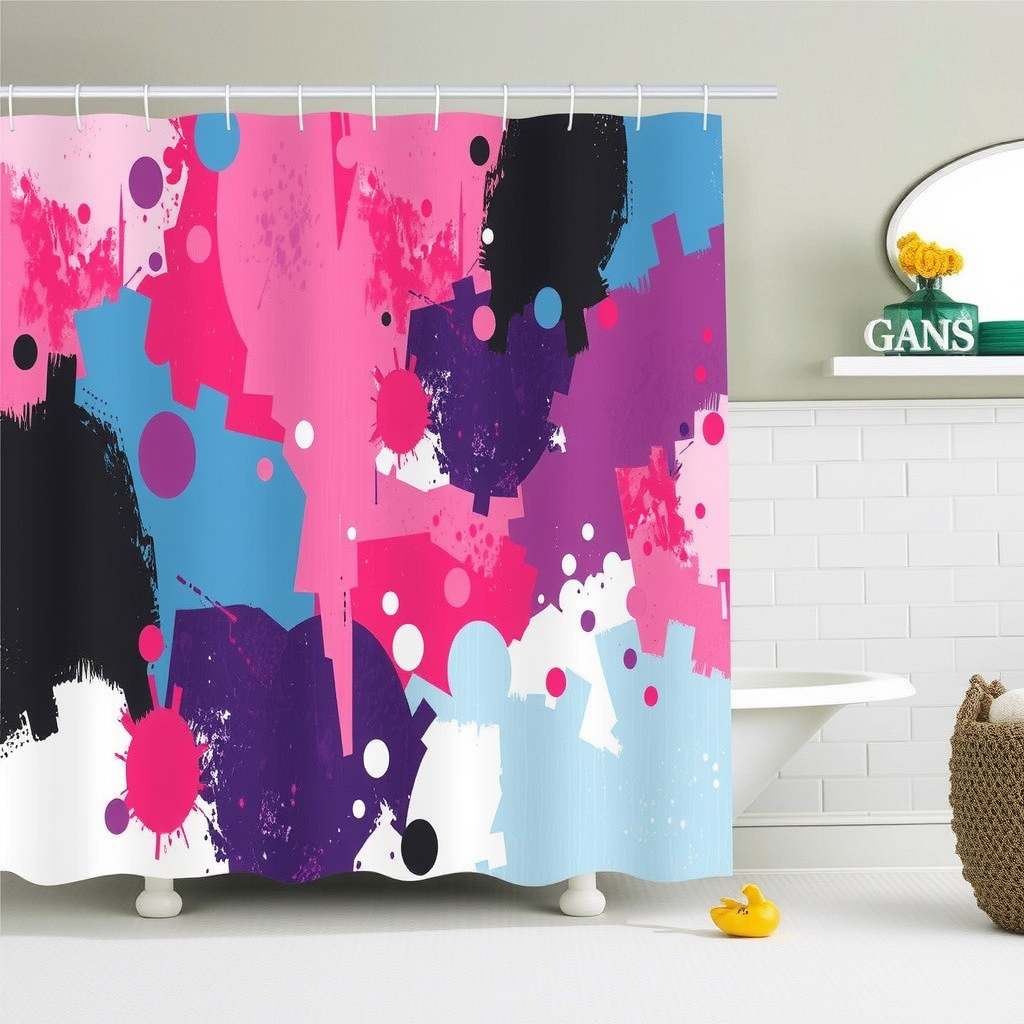} &

        \includegraphics[width=0.13\linewidth]{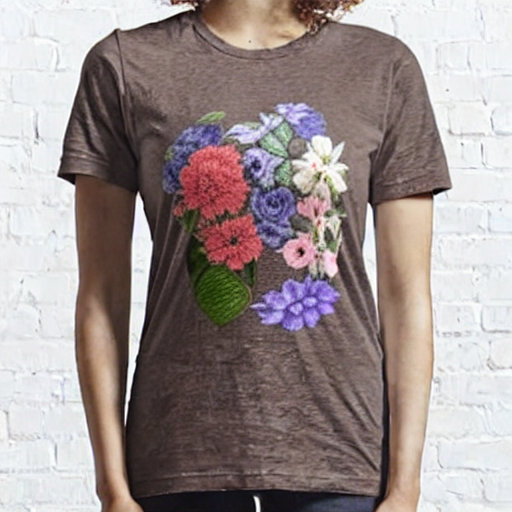} &
        \includegraphics[width=0.13\linewidth]{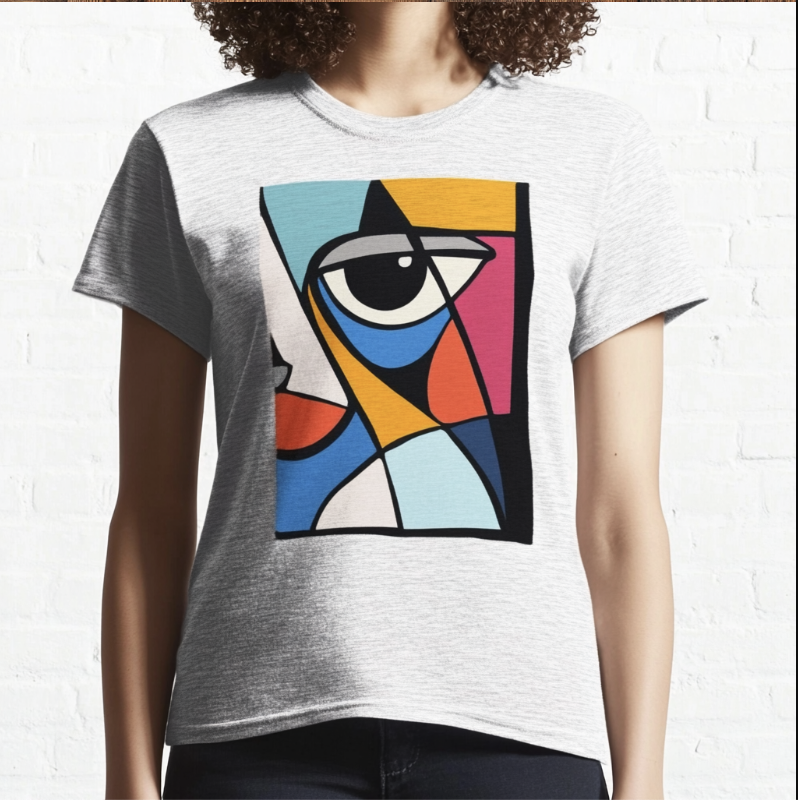} \\
    \end{tabular}

    \caption{
        Template-memorized images extracted from state-of-the-art models.
        For each model, the source image (left) and corresponding memorized generation (right) are shown.
    }
    \label{fig:sota_results}
\end{figure*}

While we focused our efforts and designed our attack on Stable Diffusion 1.4 (SD 1.4), we could also extract template-memorized images from other popular generative models. Primarily, by using the same collocations identified in SD 1.4. In \cref{fig:other_models} we exhibit further images extracted from 
\emph{DeepFloyd IF-XL-I-v1.0} \cite{DF1HuggingFace} and Midjourney V4 \cite{Midjourney}. Both models were released around the time of SD 1.4, and in \cref{fig:sota_results} we demonstrate further extractions from state-of-the-art (SOTA) models.
\begin{itemize}
\item \textbf{DeepFloyd IF-XL-I-v1.0}: although the generated images exhibited lower contrast, which made clique search impractical, we successfully identified template-memorized images through visual inspection. In addition, we discovered new templates, absent from SD 1.4, using Google Lens searches on manually flagged generations.

 \item \textbf{Midjourney V4:} MidJourney \cite{Midjourney} posed additional challenge for our attack, due to the lack of an open-source implementation or public API. However, by restricting our experiments to a limited set of prompts and seeds, selected based on template-memorized prompts identified in SD 1.4, we could still apply our attack. 
 
 For each prompt and seed, we collected the four default outputs returned by Midjourney, specifying version 4 with the argument \texttt{--v 4}. Despite limited prompt and seed control, we successfully extracted template-like generations from Midjourney. Our results included both direct template replications from SD 1.4 and interpolated reconstructions (see\cref{ssec:interpolations})). 

While the training data for Midjourney remains undisclosed, given the scale of the model and industry norms, we assume that large portions of the training corpus were sourced from publicly available internet data.
\begin{figure}[h]
    \centering
    \scriptsize

    \textbf{DeepFloyd} \\
    \setlength{\tabcolsep}{1pt} 
    \begin{tabular}{cccccc}
        \textbf{Source} & \textbf{Generated} &
        \textbf{Source} & \textbf{Generated} &
        \textbf{Source} & \textbf{Generated} \\
        \includegraphics[width=0.135\columnwidth]{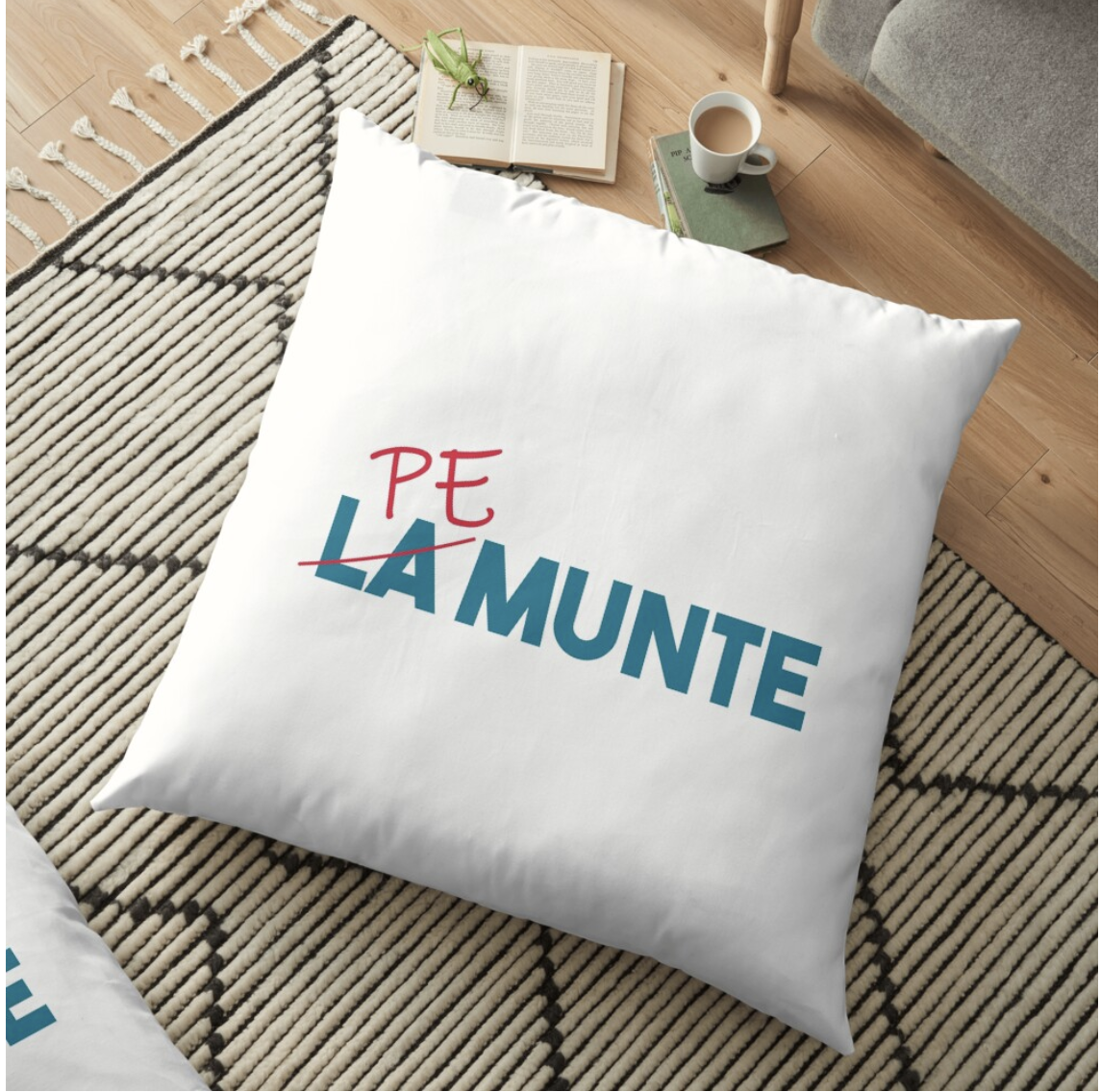} &
        \includegraphics[width=0.135\columnwidth]{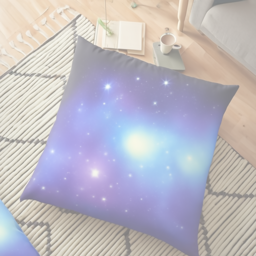} &
        \includegraphics[width=0.135\columnwidth]{main_example_images/beach_towel.jpg} &
        \includegraphics[width=0.135\columnwidth]{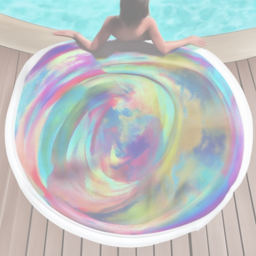} &
        \includegraphics[width=0.135\columnwidth]{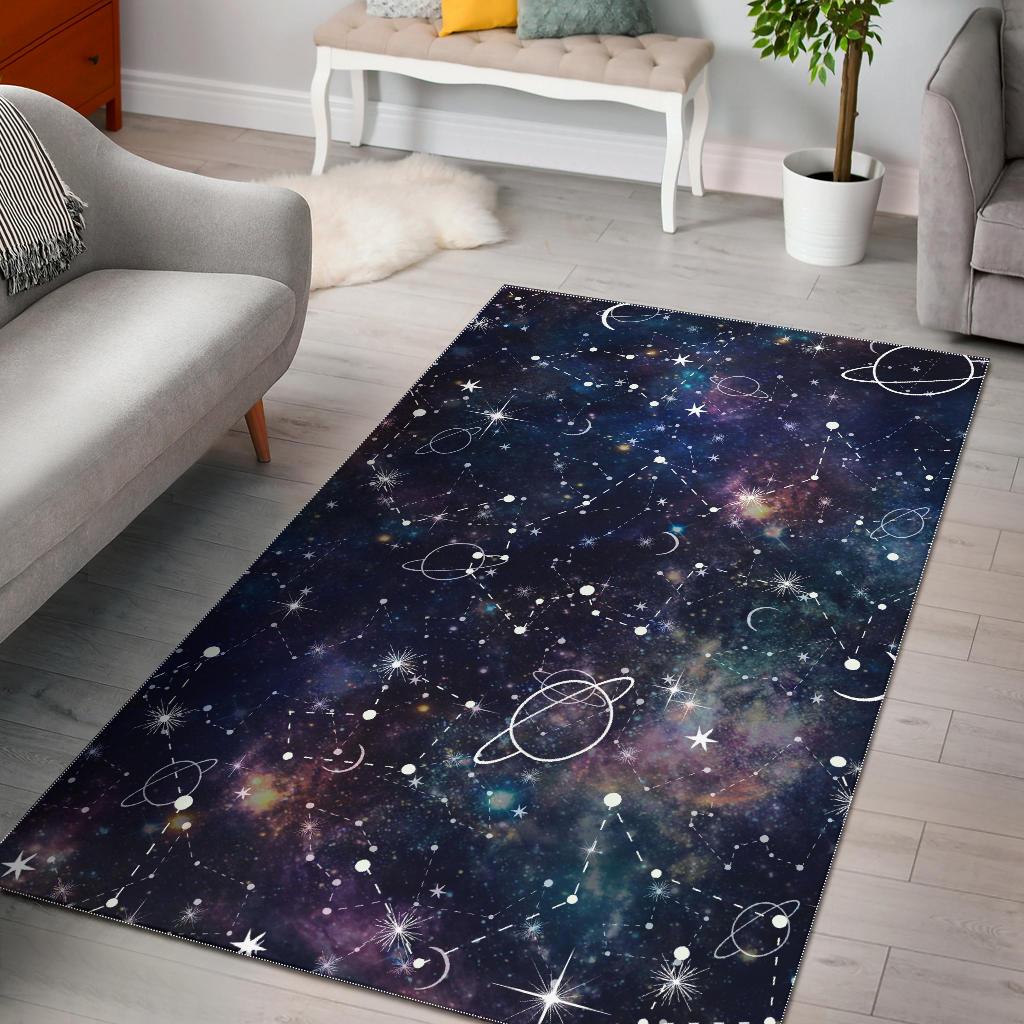} &
        \includegraphics[width=0.135\columnwidth]{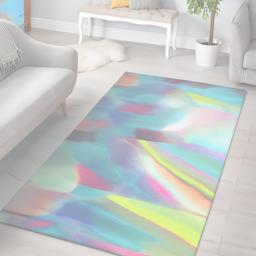} \\
    \end{tabular}
    \caption*{Example results from our attack on \textbf{DeepFloyd}. For each pair, the source image (left) and the generated image (right) are shown.}

    \vspace{0.4cm}

    \textbf{Midjourney v4} \\
    \setlength{\tabcolsep}{1pt}
    \begin{tabular}{cccccc}
        \textbf{Source} & \textbf{Generated} &
        \textbf{Source} & \textbf{Generated} &
        \textbf{Source} & \textbf{Generated} \\
        \includegraphics[width=0.135\columnwidth, height=0.135\columnwidth]{example_images/other_models/floor_pillow_source.png} &
        \includegraphics[width=0.135\columnwidth, height=0.135\columnwidth]{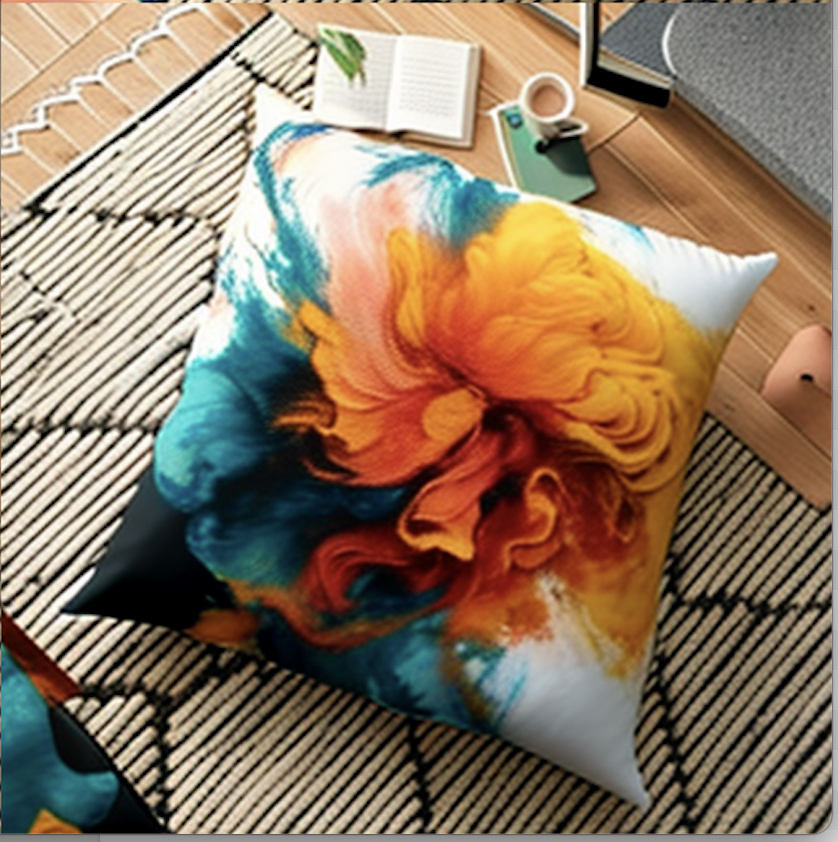} &
        \includegraphics[width=0.135\columnwidth, height=0.135\columnwidth]{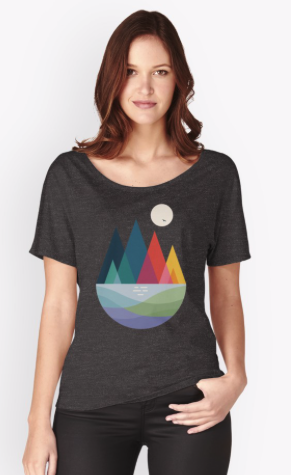} &
        \includegraphics[width=0.135\columnwidth, height=0.135\columnwidth]{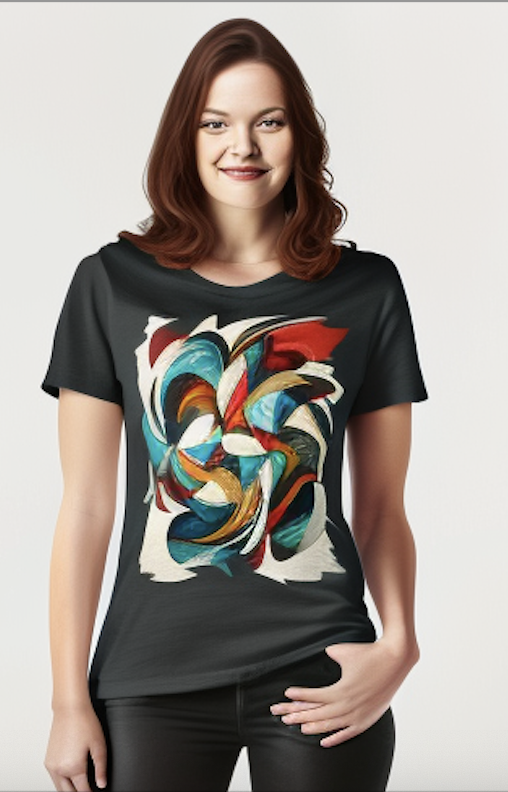} &
        \includegraphics[width=0.135\columnwidth, height=0.135\columnwidth]{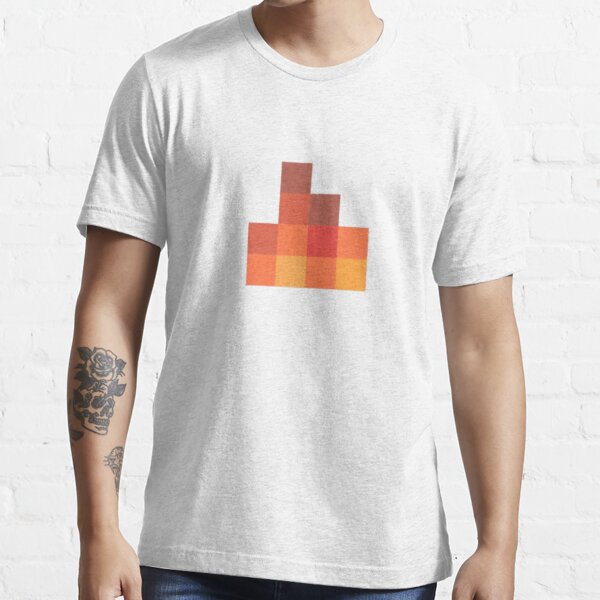} &
        \includegraphics[width=0.135\columnwidth, height=0.135\columnwidth]{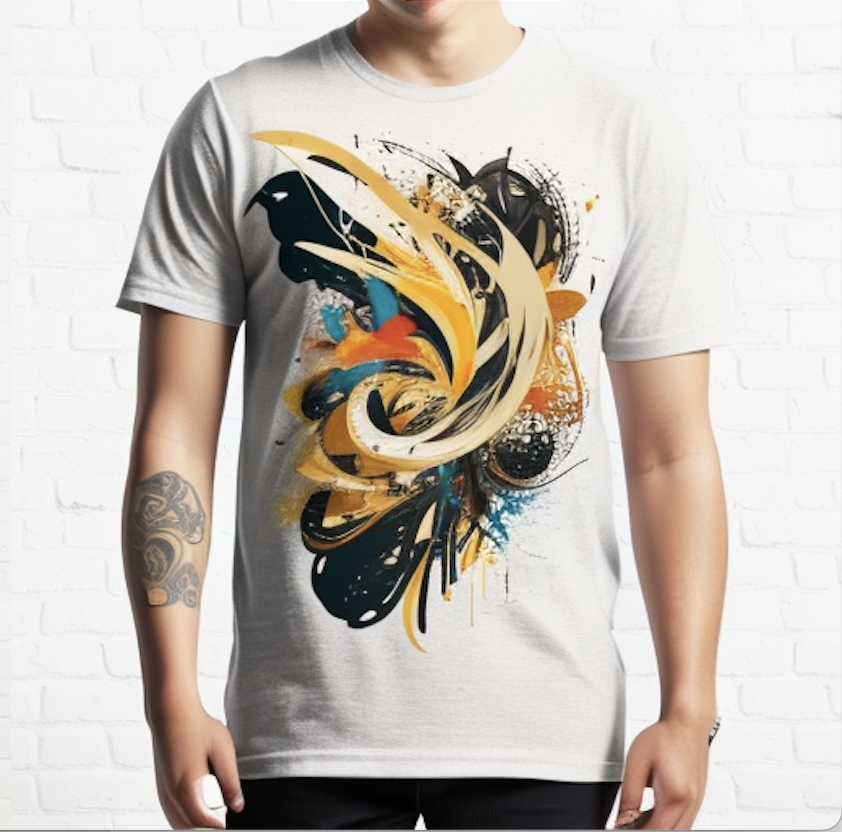} \\
    \end{tabular}
    \caption{
        Reconstructed images from models of the same generation as SD~1.4.
        For each pair, the source (left) and generated (right) images are shown.
    }
    \label{fig:other_models}
\end{figure}

\item \textbf{SOTA Models:}
In addition to attacking models from the same generation of SD, we further evaluated the vulnerability of state-of-the-art (SOTA) text-to-image models using collocations identified as template-memorized in SD 1.4. We attacked the following models: \textit{Stable Diffusion 3.5 Medium} ~\cite{SD35HuggingFace}, \textit{Flux-Schnell v1.0}~\cite{FluxSchnell}, and \textit{Midjourney v6.1} ~\cite{Midjourney}.

For Stable diffusion and flux Schnell we used the Hugging Face checkpoint in the same manner we've conducted our main experiments on SD v1.4, while for Midjourney we used the Discord interface as was detailed previously. We successfully demonstrated some memorization, as seen in \cref{fig:sota_results}, though to a lesser extent than in older models and versions. 

\textbf{SD 3.5} introduced partial mitigation techniques targeting image-text coupling vulnerabilities; however, these measures were not specifically designed to address the template-style coupling exploited by our attack, which relies on concise prompts derived from external sources rather than training data. Our results indicate that SD 3.5 exhibits increased resilience but remains partially vulnerable to template memorization. Representative examples are provided in \cref{fig:sota_results}

\textbf{MidJourney V6.1}, as discussed, poses additional challenges. However, we applied the same methodology used for V4. While the generated images were typically perturbed (see (\cref{ssec:perturbations}), targeting previously vulnerable benign prompts still yielded images with notable resemblance to the source. For instance, as shown in \cref{fig:sota_results}, the prompt ``Abstract Art Essential T-Shirt" (which previously produced near-verbatim copies) continues to generate closely resembling images. 
\end{itemize}

\subsection{User Study}\label{ssec:user_study}
To validate our manual annotations of copied images, we conducted a user study, to confirm that the copying is visually apparent to a human observer. Prior work focused on verbatim copies, where metrics like SSCD or edge detection were incorporated in the pipeline. In contrast, our work targets template-level memorization, where the reproduced content may be in the background, or differ in minor structure. This makes the problem of copying ill-posed, and we observed that edge consistency frequently failed to flag images that, to a human observer, were clearly copied. We therefore validated our observations through a user study that confirmed the success of our attack. Participants were shown generated images and potential source images, and were asked to identify copied images. The selected examples included both direct instances of TMI and cases of interpolations and perturbations, as described in \cref{ssec:interpolations} and \cref{ssec:perturbations}, respectively. Overall, we collected a dataset of 56 source images and corresponding generated counterparts. 70 users answered our questionnaire and each user had to answer 18 to 19 questions.  For each generated image, we also selected a control image which was another output produced using the same prompt but a different, randomly chosen seed. We also showed examples where \textbf{both} generated images and \textbf{none} contained copied elements from the source. In each question, participants were shown a \textbf{triplet} consisting of a source image and two generated images, and were asked:

\begin{quote}
\textit{``For each image pair, select which generated image appears to have elements copied from the source.''}
\end{quote}

\begin{figure}[h]
    \centering
    \includegraphics[width=0.5\linewidth]{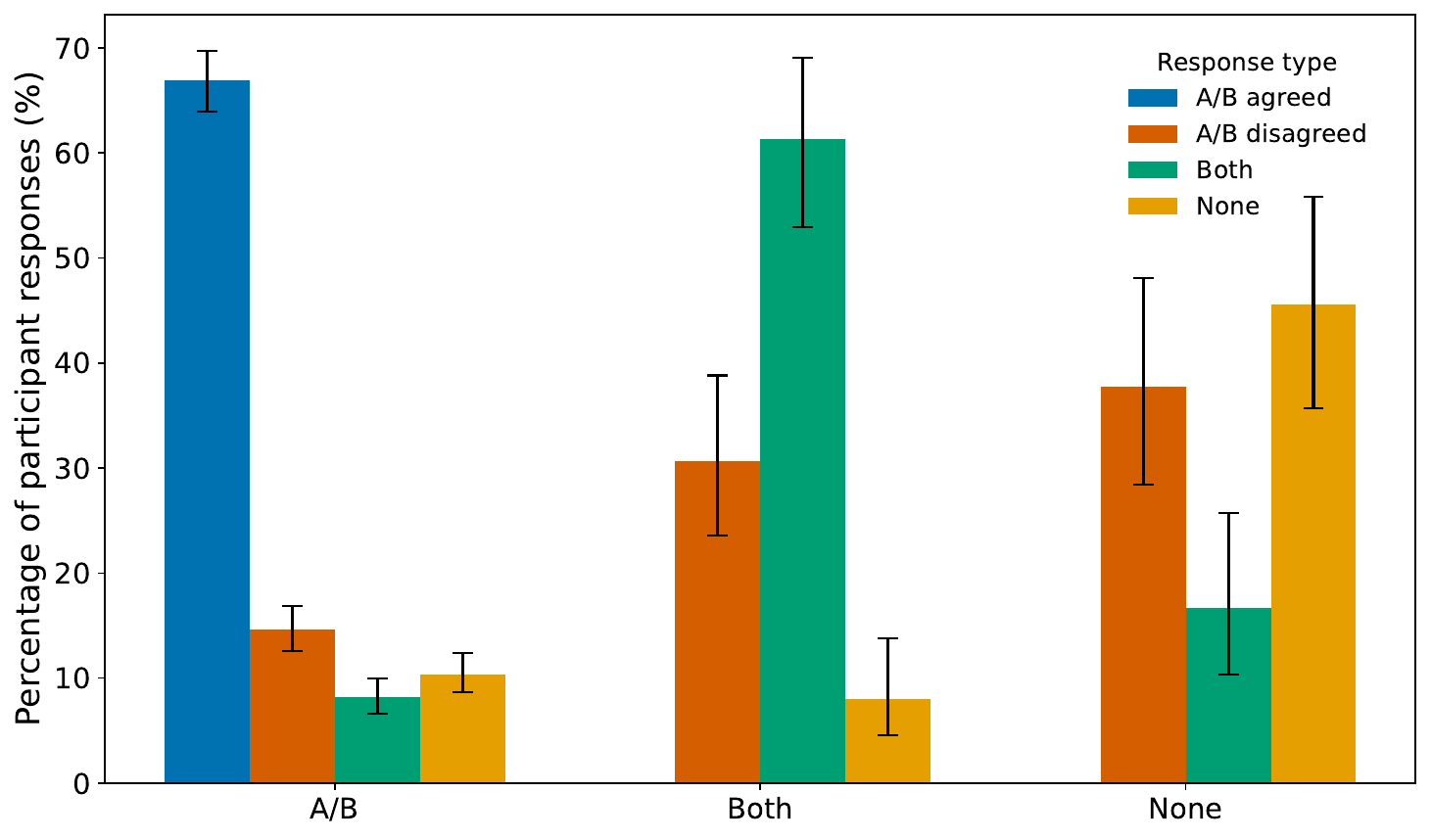}
    \caption{User study results aggregated by our ground-truth labels with cross-participant CI.}
    \label{fig:user_study_results}
\end{figure}

Figure~\ref{fig:user_study_results} presents a summary of the study, showing that user judgments validate our identification of copied content.

The available choices were: A, B, Both, and None. The questionnaires were administered through Google Forms, and the participants non-expert volunteers recruited through personal networks. To verify the diversity of the sample, we also collected age and gender information.

\begin{figure}[h]
    \centering
    \includegraphics[width=0.5\linewidth]{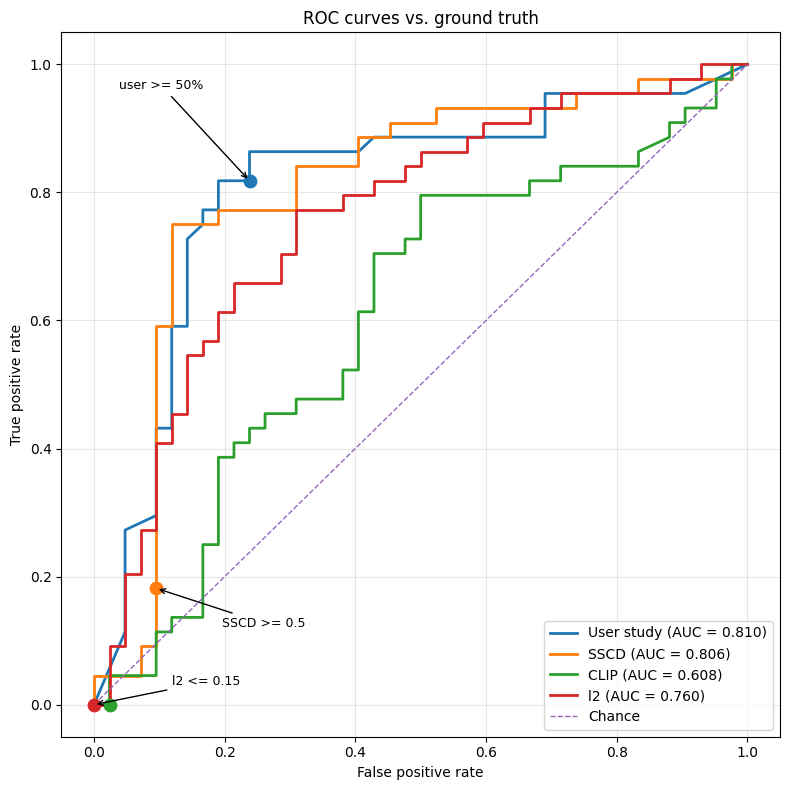}
    \caption{Comparison of similarity metrics with the user study, the thresholds marked on the graph are user study majority vote, SSCD from \cite{Somepalli2023}, and $\ell_2$ from \cite{carlini2023}}
    \label{fig:metrics_vs_user_study_roc}
\end{figure}

We calculated the pairwise similarity between each image and the source image $s(source, TMI), s(source, control)$ used in previous works and compared them to the percentage of users who identified the image as having elements copied from the source. The results are shown in \cref{fig:metrics_vs_user_study_roc}. 
 It is important to notice that each pair the image and the source were generated from the same prompt, which biases the similarities upward.

\subsection{Real-World Benign Prompts}\label{ssec:prompts_in_the_wild}
We previously characterized our attack as reflecting the behavior of a naive user who interacts with the model exactly as intended. To further support this claim, we analyze two large-scale datasets of real user prompts \cite{MJprompts, SDprompts}, collected directly from everyday model usage.

First, in each dataset, we search for prompts containing our identified collocations to validate that indeed such collocations appear in every-day use. In the Midjourney dataset \cite{MJprompts}, we managed to find 12,706 total prompts (3,883 unique) containing these collocations out of 7.13M samples. In the Stable Diffusion dataset \cite{SDprompts}, we could find 818 total prompts (660 unique) out of 1.8M samples.

The original images generated from these prompts were unavailable due to link expiration (Midjourney) and dataset scale (DiffusionDB) so we could not estimate how many of these prompts actually yielded TMIs in real-world use cases. Nevertheless, as a preliminary assessment, we conducted a small, low-resource experiment focusing on a limited set of selected candidate prompts From the 3,883 unique Midjourney prompts and 660 unique DiffusionDB prompts that contain our collocations, we select only the shortest prompt, yielding curated subsets of 35 candidate prompts. We then generated images using our previously described pipeline and manually inspect them for the templates we have identified (see \ref{apx:template_groups}). Despite the modest scale of this experiment, the results were notable: 8 out of 20 Midjourney prompts and 7 out of 15 DiffusionDB prompts produced TMIs. An example prompt and its resulting TMI appear in \cref{fig:wild_prompt_example}. Although preliminary, these findings suggest that TMIs may be more common in the wild than previously assumed. A more extensive study is needed to determine their true prevalence and evaluate the seriousness of this phenomenon.

\subsection{Comparison with previous works}
\begin{figure}[h]
\centering
\scriptsize
\setlength{\tabcolsep}{2pt}
\renewcommand{\arraystretch}{1.2}

\begin{tabular}{
  >{\centering\arraybackslash}m{0.25\columnwidth}
  >{\centering\arraybackslash}m{0.25\columnwidth}
  >{\raggedright\arraybackslash}m{0.45\columnwidth}
}
\toprule
\textbf{Generated Image} & \textbf{Source Image} & \textbf{Prompt and Source Caption} \\
\midrule

\includegraphics[width=0.8\linewidth]{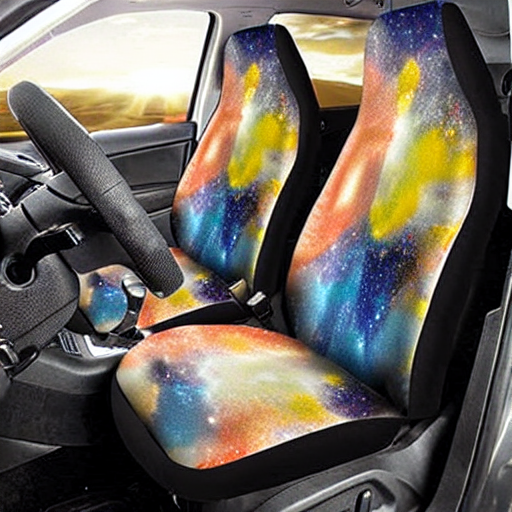} &
\includegraphics[width=0.8\linewidth]{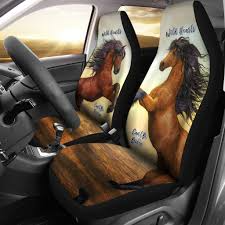} &
\textbf{Prompt:} Galaxy Print Universal Fit Car Seat Covers.  
\newline \textbf{Caption:} Wild Hearts Can’t Be Broken Car Seat Covers For Horse Lovers 170804 – YourCarButBetter. \\

\midrule

\includegraphics[width=0.8\linewidth]{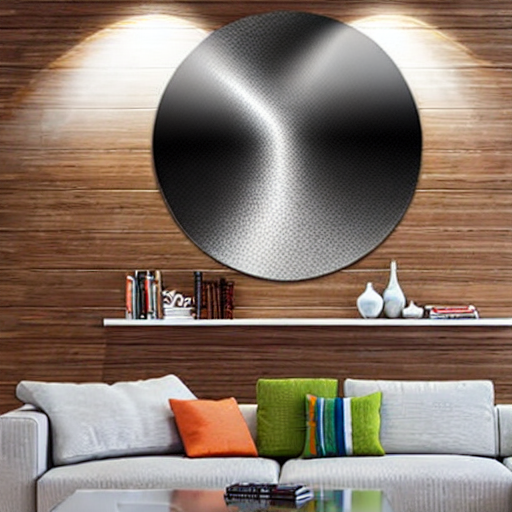} &
\includegraphics[width=0.8\linewidth]{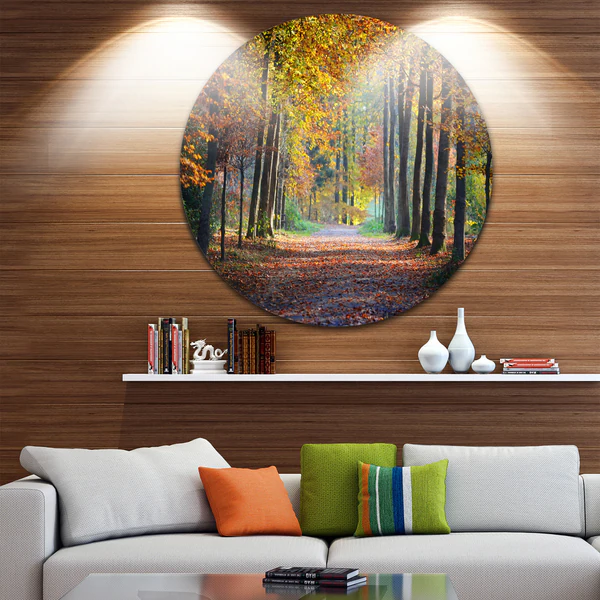} &
\textbf{Prompt:} Abstract Art Round Metal Wall Art.  
\newline \textbf{Caption:} Designart “Wide Pathway in Yellow Fall Forest” Landscape Photo Round Metal Wall Art. \\

\bottomrule
\end{tabular}
\caption{\small Comparison of Generated and Source Images with Corresponding Prompts and Captions when taking the categories from previous works, examples from \cite{hintersdorf24nemo}.}
\label{fig:benign_vs_full_prompts}
\end{figure}

As discussed, to allow comparison with previous attacks we also extracted a list of collocations, similar to the one we extracted from the generic websites we chose, by harvesting e-commerce sites that appeared in previous works. For example, we could extract the collocation \emph{car seat cover} and compare our approach to previous approach that mines captions from the training set. \cref{fig:benign_vs_full_prompts} provide examples of images and the associated prompts used in each attack.

\subsection{Traces of Memorization in Stable Diffusion 3.5}\label{apx:sd35}

Beyond the clear instances of template extraction—sometimes preserved only up to perturbations, as shown in \cref{fig:sota_results}—we also found that Stable Diffusion 3.5 exhibits subtler forms of memorization. Even when our original attack no longer yielded a recognizable reconstruction, meaningful traces of the underlying training content persisted, suggesting that future, more targeted attacks may still uncover sensitive material.

These findings highlight the significance of the phenomena described in this work—interpolations, perturbations, and template leakage. Without understanding these behaviors, one might incorrectly assume that failed reconstructions imply the absence of memorization. Instead, these phenomena reveal that memorized content can manifest in isolated visual elements, partially altered fragments, or blended structures rather than as a direct or verbatim template. In other words, the form in which memorization appears can change—even when its presence does not.

\begin{figure}[h]
    \centering
    \setlength{\tabcolsep}{0.5em} 
    \renewcommand{\arraystretch}{1.0}
    \begin{tabular}{ccc|c}
        \includegraphics[width=0.22\textwidth]{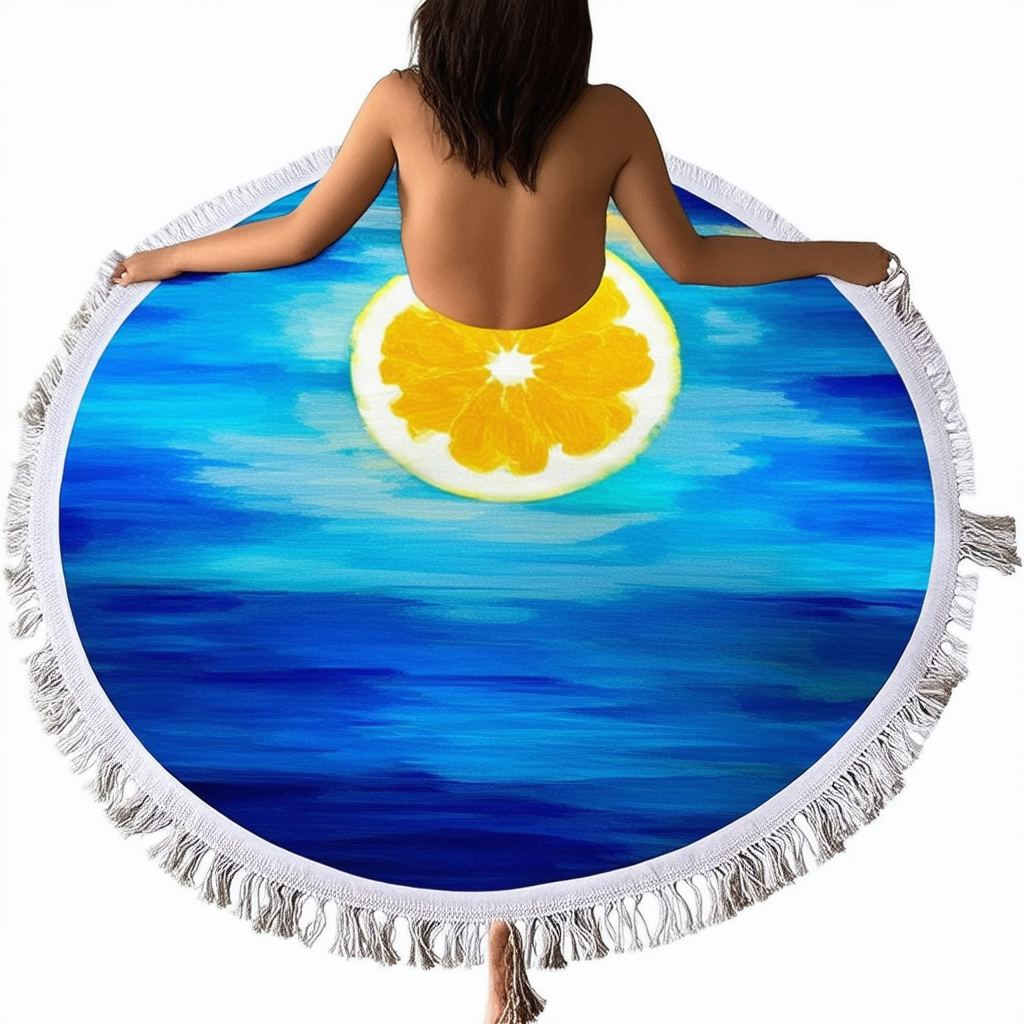} &
        \includegraphics[width=0.22\textwidth]{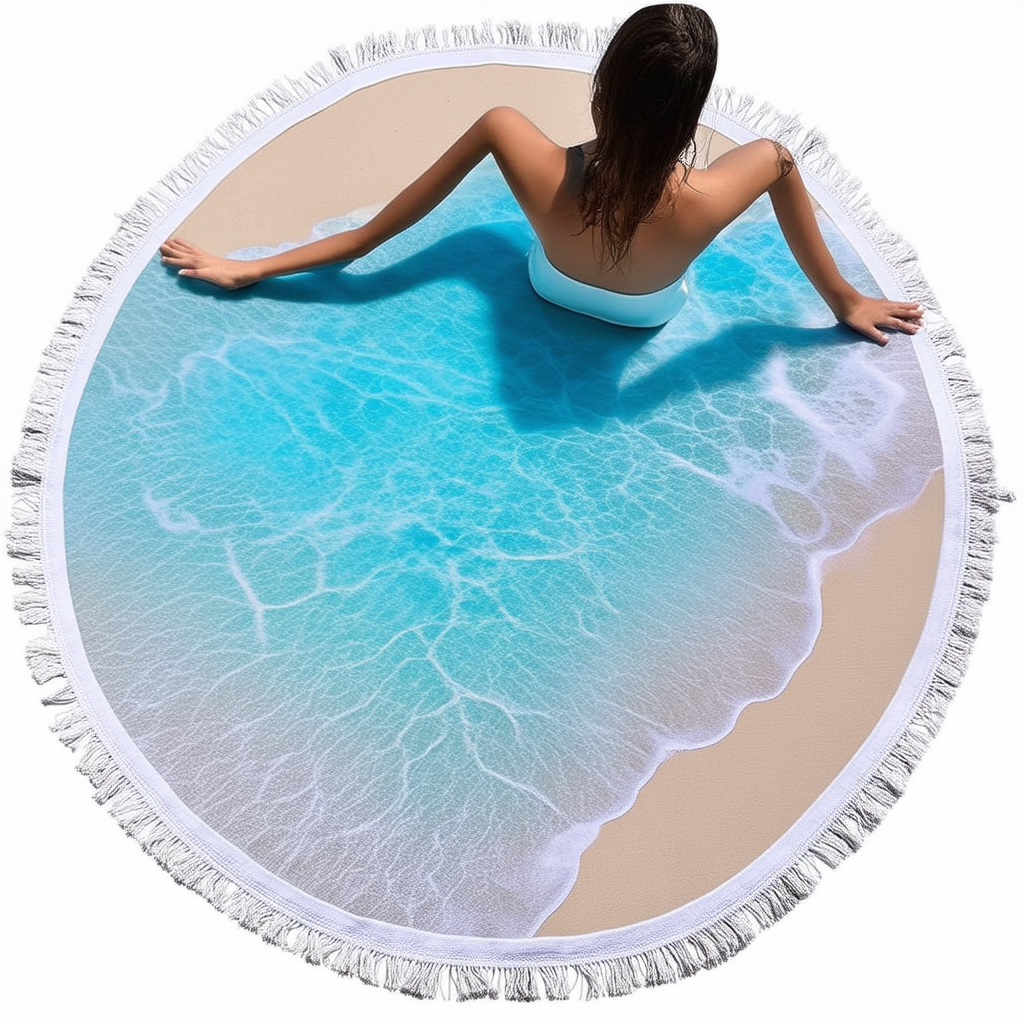} &
        \includegraphics[width=0.22\textwidth]{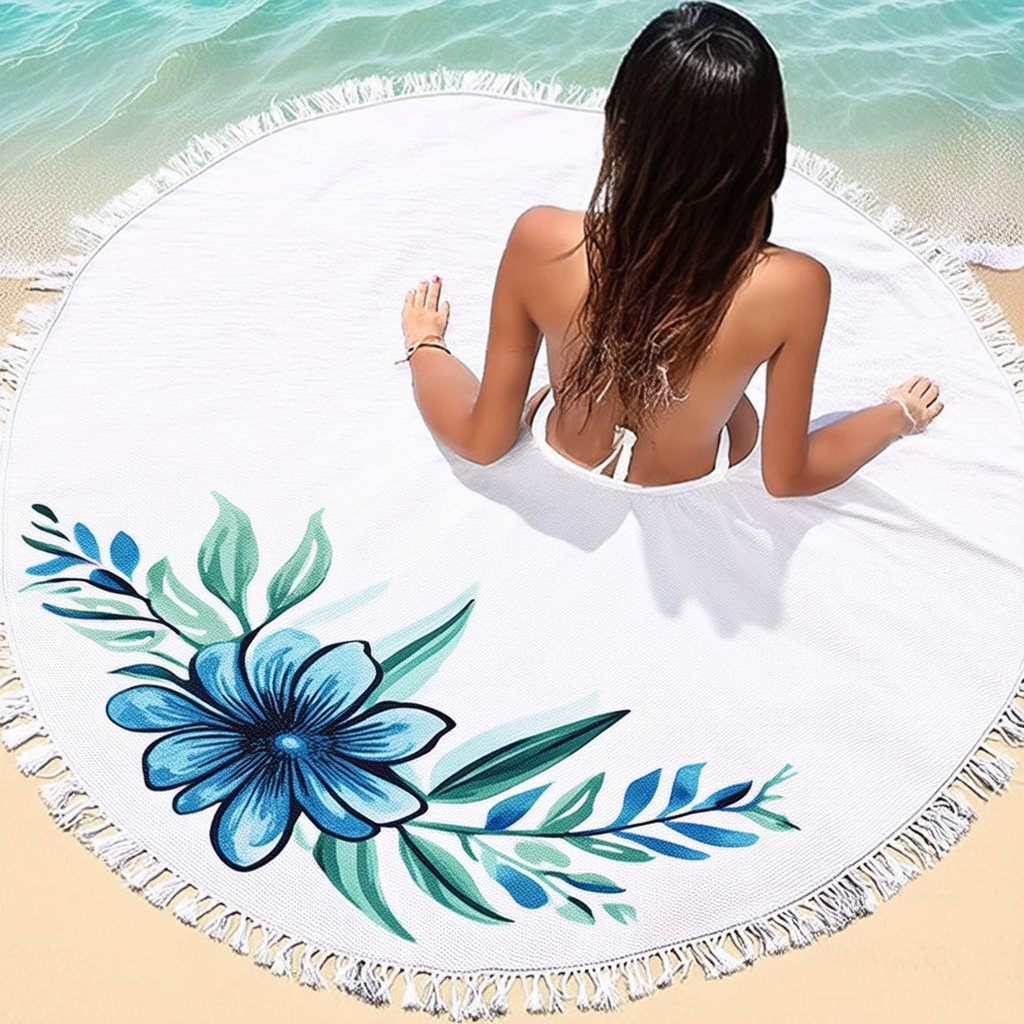} &
        \includegraphics[width=0.22\textwidth]{main_example_images/beach_towel.jpg} \\
    \end{tabular}
    \caption{Traces of our attack in SD 3.5 (Medium). The first three images were generated using the prompt \emph{"BOYOUTH Round Beach Towel,X Beach Mat with Tassels Ultra Soft Super Water Absorbent Multi-Purpose Towel,59 inch-Diameter"} Taken from the source image's product description in Amazon, where X was: "Abstract Art", "Floral", "Galaxy". The image to the right of the vertical line is the source image, which was reconstructed in previous models (see \cref{fig:generic})}
    \label{fig:sd35_beach_towel}
\end{figure}

A concrete example appears in \cref{fig:sd35_beach_towel}. Although the image template had no longer been fully reconstructed in Stable Diffusion 3.5, its generation still exhibited recognizable elements, such as a person in a specific pose. Using the template groups we had previously identified for Stable Diffusion 1.4, we were able to match this distorted output back to the same underlying template and corresponding source image. This shows that, even when verbatim extraction is no longer possible, remnants of training images, particularly from large-scale e-commerce data,can survive as structured traces tied to known templates, and thus may continue to pose a risk in future model versions.

\subsection{Synthetic Experiments}
To better understand how the coupling between \emph{text templates} and \emph{image templates} contributes to partial memorization, we designed a series of synthetic experiments that deliberately recreate this form of coupling. Our goal was twofold: (1) to intentionally induce memorization under controlled conditions, and (2) to investigate another phenomenon we call \emph{template leakage}. In real-world data, we frequently encountered images where a template from one category appeared under a different category. However, because e-commerce platforms often reuse templates across product types, it was impossible to determine whether these cases reflected true leakage or simply the natural reuse of templates in the training data. Synthetic experiments offered a way to isolate these effects.

\paragraph{Stage 1: Controlled overlays with crude template replacement.}
For the first stage, we collected three photographs of a coffee mug next to an iPhone SE, placing the same mug in three different locations in our lab. We then created manual masks for each image using Photopea and replaced the masked regions with simple patterns using OpenCV. At this stage, the overlays were intentionally crude---visually unrealistic and easy for the model to memorize.

Under these conditions, we observed \emph{verbatim template extraction}, as expected, along with clear instances of \emph{interpolation}, \emph{perturbations}, and \emph{leakage}. The model reproduced not only the underlying structure of the templates but also combined and varied them in ways that matched the phenomena we identified in real-world data.

\paragraph{Stage 2: Realistic mockups mimicking Print-on-Demand rendering.}
In the second stage, we constructed a more realistic setup based on Print-on-Demand (PoD) workflows. We selected three mockups from Freepik and replaced their smart-object contents in Photoshop using an automation script, with light manual adjustments to maintain uniformity across images. For one mockup, we additionally inserted two decorative elements---a pair of leaves and a lemon slice---also from Freepik, to mimic the kind of compositional variability seen in real product designs.

With these more naturalistic overlays, we no longer observed verbatim template extraction. However, \emph{interpolation}, \emph{perturbations}, and \emph{object memorization} remained evident, confirming that these behaviors persist even when the templates are more realistically embedded and less trivially memorized.

\paragraph{Testing for template leakage under semantic shifts.}
To probe template leakage directly, we generated images not only for the collocation ``Coffee Mug'' but also for prompts that were \emph{semantically similar} (``Tea Cup'') and \emph{semantically dissimilar} (``T-Shirt''). These generations allowed us to separate genuine leakage from simple template reuse.

The results provided clear evidence of genuine leakage. Under prompts such as ``skg Tea Cup,'' the backgrounds consistently showed interpolations---or, at minimum, strong color and texture echoes---of the coffee-mug templates used during fine-tuning. Even more strikingly, the generated tea cups appeared in \emph{top view}, a viewpoint that never appeared in the training data and was absent from all coffee-mug generations, which only reproduced the three canonical views present in the training set. The cups also manifested in shapes not seen during training.

Together, these findings demonstrate not only that template leakage occurs, but also that synthetic experiments enable us to disentangle leakage from benign template reuse- revealing how template structures can transfer across prompts and categories under controlled conditions.

\begin{figure}[h]
\begin{tabular}{>{\centering\arraybackslash}m{0.15\linewidth} 
                >{\centering\arraybackslash}m{0.38\linewidth} 
                >{\centering\arraybackslash}m{0.30\linewidth}}

  & stage I & Stage II\\
Training data &  
\includegraphics[width=0.4\textwidth]{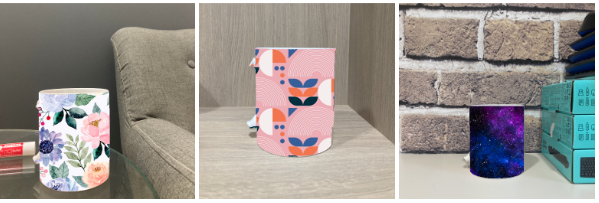} &
\includegraphics[width=0.4\textwidth]{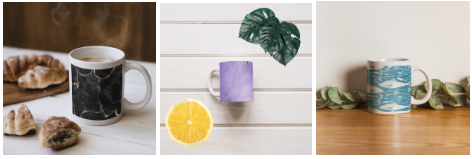}\\
"Coffee Mug" &
\includegraphics[width=0.4\textwidth]{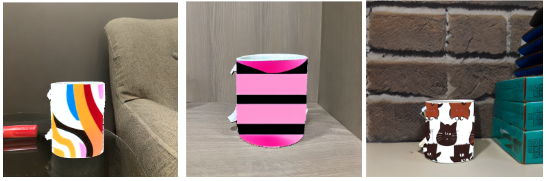} &
\includegraphics[width=0.4\textwidth]{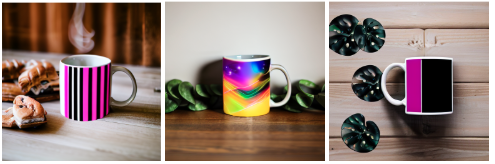}\\
"Tea Cup" &
\includegraphics[width=0.4\textwidth]{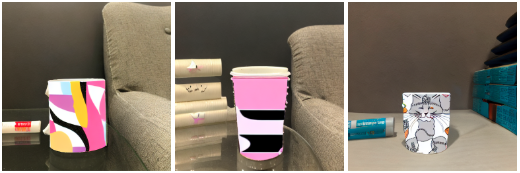} &
\includegraphics[width=0.4\textwidth]{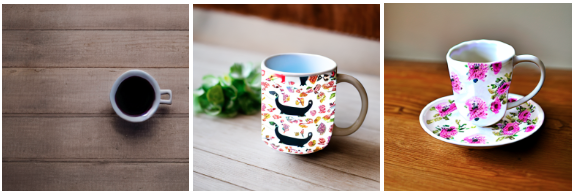}\\
"T-Shirt" &
\includegraphics[width=0.4\textwidth]{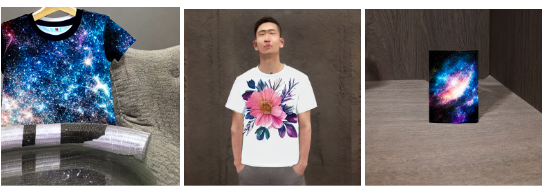} &
\includegraphics[width=0.4\textwidth]{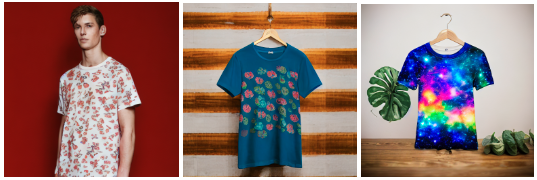}\\

\end{tabular}
 \caption{Intentionally causing template memorization by fine tuning SD on coupled image-text pairs. The generated results demonstrate the phenomena of interpolation, perturbations, and leakage.}
    \label{tab:LoRa results}
\end{figure}

\subsection{Evaluation of BE-PRSS Detection and Mitigation}
\label{sec:beprss_eval}

We evaluated the detection and mitigation methods proposed by Chen et al. \cite{chen2025exploringlocalmemorizationdiffusion} on our template-memorized examples. We constructed prompts by adding the descriptor ``Abstract Art'' to each tested collocation, and ran each prompt with $3$ different seeds using the BE-PRSS detection script. We manually annotated each generated image as TMI or non-memorized by comparing it to the corresponding known template. Some outputs annotated as TMI were interpolation cases, as described in Section~\ref{ssec:interpolations}.

Using the default BE-PRSS threshold of $1.0$, the detector reached an accuracy of $0.595$, precision of $0.520$, and recall of $0.722$. Thus, while the method detected some TMI cases, it was not reliable for template-level memorization in our setting.

We also tested the mitigation method on the $43$ collocations previously associated with memorization. For each collocation, we generated $25$ prompt variations using ChatGPT, as required by the mitigation script, and counted whether at least one of three seeds produced a TMI. Before mitigation, $28$ collocations produced a TMI; after mitigation, $22$ did. This suggests only a modest reduction. We also observed utility degradation, where some mitigated prompts no longer generated the requested item. We believe this is due to the method's reliance on prompt variations, which assumes memorization is tied to a specific prompt, whereas our results suggest a many-to-many structure driven by collocations or keywords rather than the full prompt.
\section{Observed Phenomena in Template Memorized Images}\label{apx:observed_phenomena}
Building on the successful reconstruction of a large set of template-memorized images, we observed three notable phenomena—\textbf{Interpolation}, \textbf{Perturbations}, and \textbf{Leakage}—which, to the best of our knowledge, have not been previously reported in the literature.

Before moving to describe the phenomena, we first note that the generation of template-memorized content follows a many-to-many structure which we name \textbf{template groups}: a set of collocations are associated with a set of image templates, rather than the one-to-one text-image association typically assumed. This structure was hinted at \cite{webster2023} but not named and studied. We further note that the collocations in the group are either semantically similar (e.g. duvet cover, comforter, throw blanket) or appeared together in product description, as they were listed together as options of the same product (e.g. picnic blanket, beach towel).
The template groups we identified are listed at \ref{apx:template_groups}.

\subsection{Interpolation} \label{ssec:interpolations}
\begin{figure}[H]
\begin{center}
\renewcommand{\arraystretch}{1.9}
\setlength{\tabcolsep}{3pt}

\begin{tabular}{
    >{\centering\arraybackslash}m{2.8cm} |
    >{\centering\arraybackslash}m{2.4cm}
    >{\centering\arraybackslash}m{2.4cm} ||
    >{\centering\arraybackslash}m{2.4cm}
    >{\centering\arraybackslash}m{2.4cm}
}
\textbf{Source Images} &
\includegraphics[width=\linewidth]{example_images/gen_vs_source/essential_t-shirt_men.jpg} &
\includegraphics[width=\linewidth]{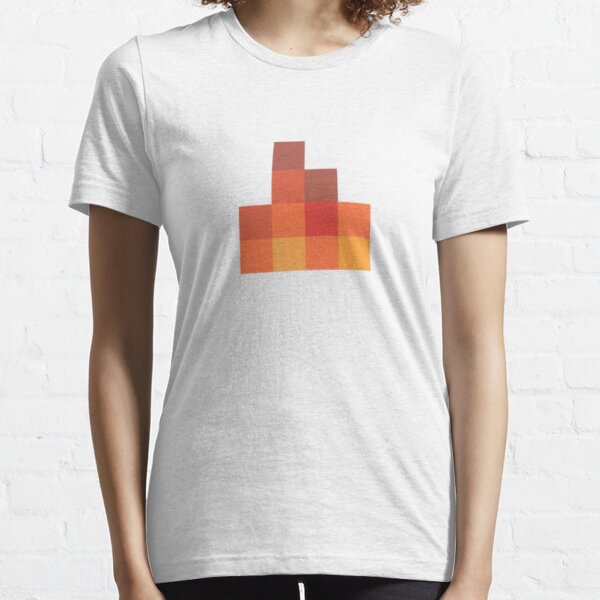} &
\includegraphics[width=\linewidth]{example_images/interpolation_examples/clear_case_for_samsung_3.png} &
\includegraphics[width=\linewidth]{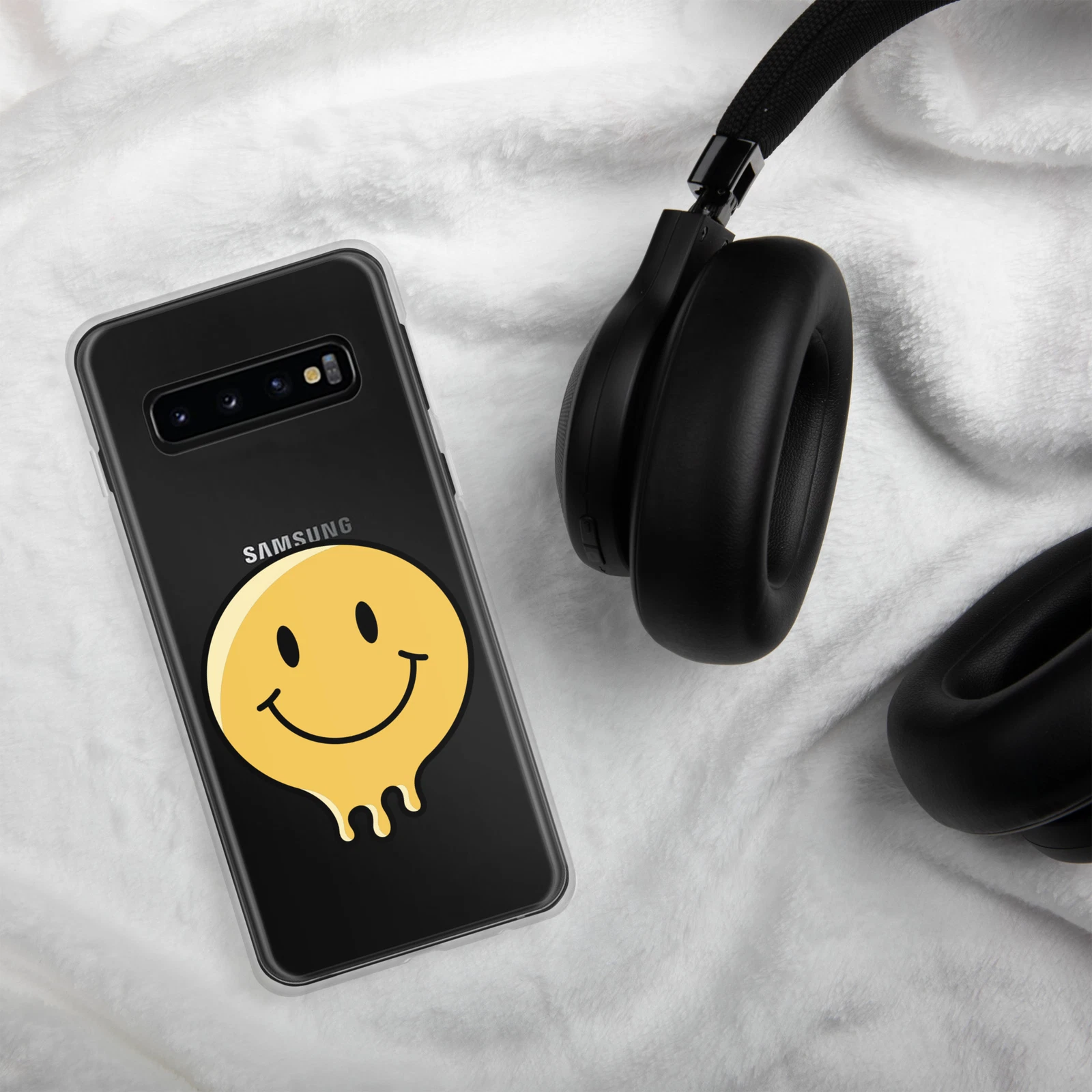} \\
\midrule
\textbf{Generated templates} &
\includegraphics[width=\linewidth]{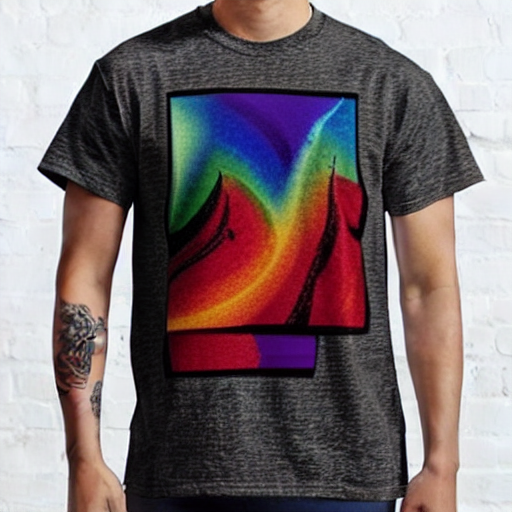} &
\includegraphics[width=\linewidth]{example_images/interpolation_examples/origin2.png} &
\includegraphics[width=\linewidth]{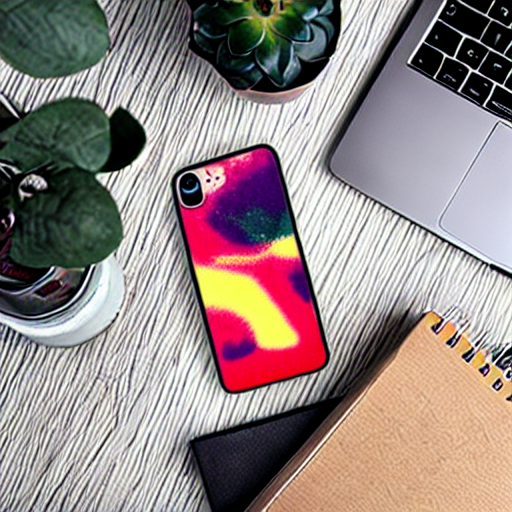} &
\includegraphics[width=\linewidth]{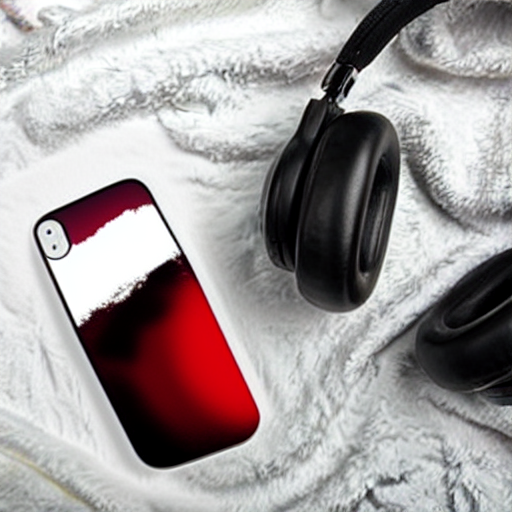} \\
\textbf{Interpolations} &
\includegraphics[width=\linewidth]{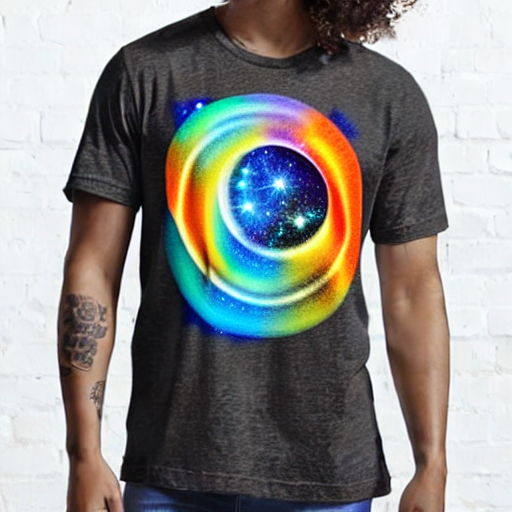} &
\includegraphics[width=\linewidth]{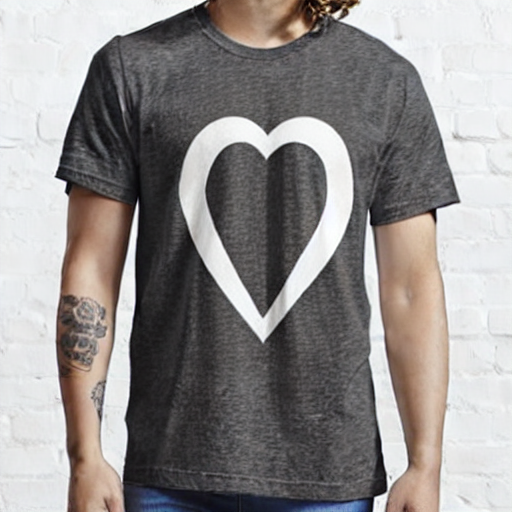} &
\includegraphics[width=\linewidth]{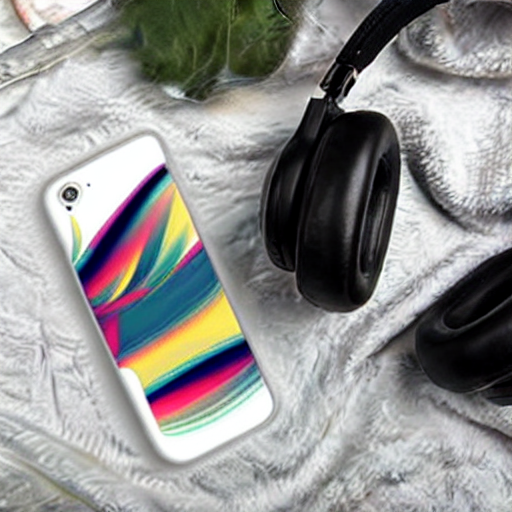} &
\includegraphics[width=\linewidth]{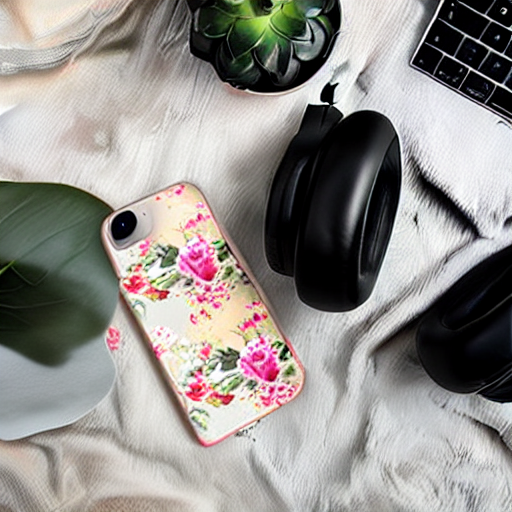} \\
\end{tabular}

\caption{Examples of interpolations we observed, left: "Essential T-Shirt", right: "iPhone Case \& Cover".}
\label{tab:interpolation}
\end{center}
\end{figure}

The first phenomena we observe involves interpolated reconstructions, where generated images blend elements from multiple training examples rather than copying any single image outright. Such partial copying evades standard similarity metrics, as neither the background nor foreground is directly duplicated, but the elements can be easily identified as matching each source by a human observer, as further validation in the user study presented in \ref{ssec:user_study}. In \cref{tab:interpolation}, for instance, one image contains a tattoo reproduced from a real-world photo, while the hair reproduced from another. The combination of elements from multiple sources raises challenging questions regarding what constitutes a memorized output.

This form of interpolation reveals a deeper behavior: diffusion models appear to internalize, reorganize, and recombine fine-grained visual traits drawn from multiple training images. In our experiments, we were able to identify interpolated reconstructions that clearly fused elements from two—sometimes three—distinct memorized sources.

We further speculate that some generations may blend far more subtle fragments from a broader set of images, producing composites whose lineage spans numerous sources. If such multi-source blending indeed occurs, it becomes increasingly difficult—sometimes effectively impossible—to determine which training images influenced a given output. This, in turn, raises philosophical questions about whether these models can be said to generate truly novel samples, or whether every output remains, in part, a rearrangement of prior data.

\subsection{Perturbations} \label{ssec:perturbations}

\begin{figure}[H]
    \centering
    \setlength{\tabcolsep}{0.5em} 
    \renewcommand{\arraystretch}{1.0}
    \begin{tabular}{llll||c}
        \includegraphics[width=0.18\textwidth]{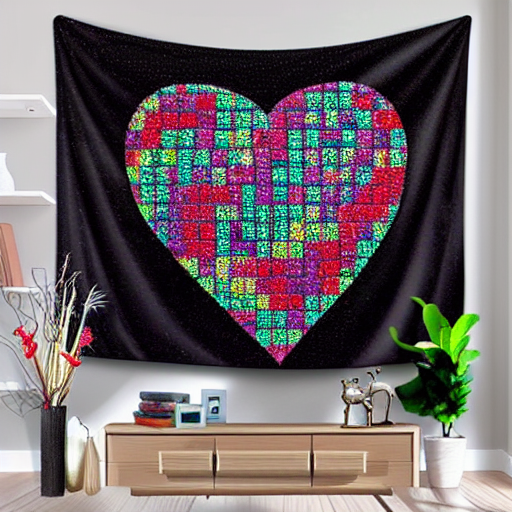} &
        \includegraphics[width=0.18\textwidth]{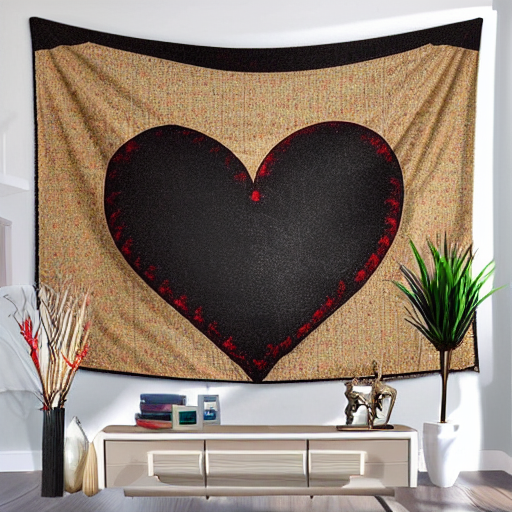} &
        \includegraphics[width=0.18\textwidth]{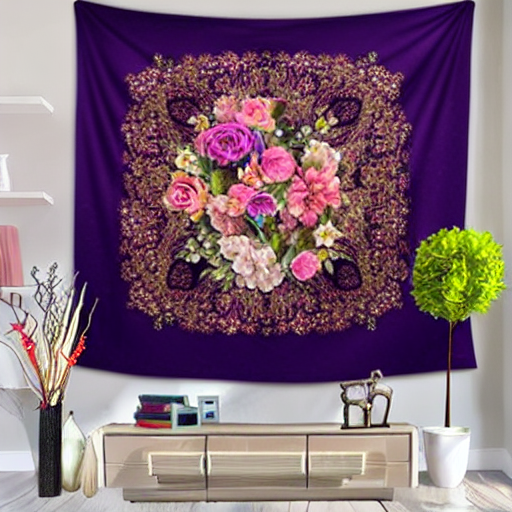} &
        \includegraphics[width=0.18\textwidth]{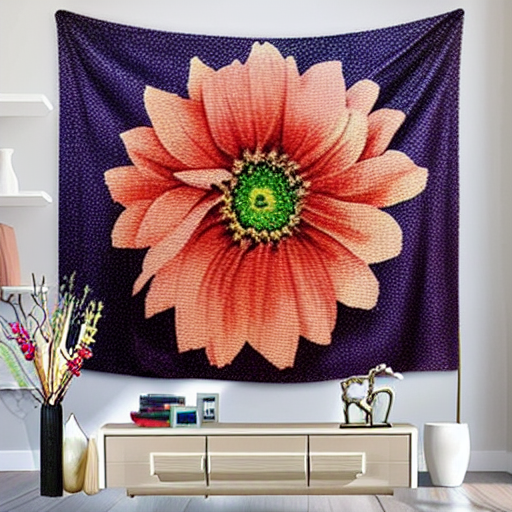} &
        \includegraphics[width=0.18\textwidth]{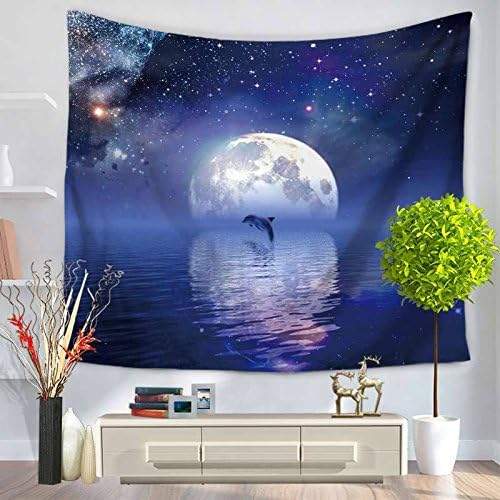}
    \end{tabular}
    \caption{Four images with perturbations (see plant on the right) generated from the prompt: “X Wall Tapestry”: where X is: "I Heart ML", "Floral". Rightmost is source found via Google Lens.}
    \label{fig:perturbations}
\end{figure}

We also observed clusters of images that were nearly identical in their underlying template, yet differed by the manifestation of one or more objects with semantically similar alternatives placed in the same location—for example, swapping a lamp for a slightly different lamp, or a chair for another chair. An illustration of this behavior appears in \cref{fig:perturbations}. These perturbations preserve high CLIP similarity on the fixed region, while introducing small pixel-level differences that modestly reduce the $\ell_2$ similarity—though the images still remain far more similar than random pairs.

The precise origin of this phenomenon is difficult to isolate, but it aligns with two plausible explanations. First, minor perturbations to the latent noise can produce small, localized variations in the final image, consistent with behavior demonstrated in \cite{kerastutorial}. Second, it may reflect the one-step denoising behavior described by \cite{webster2023}, which we observe specifically in the context of memorized images: once the global layout is recovered in the first step, subsequent steps allow only limited variation, resulting in small but consistent perturbations across generations.

\subsection{Leakage}\label{ssec:leakage}
\begin{figure}[H]
    \centering

    \begin{subfigure}{0.48\linewidth}
        \centering
        \vspace{4pt} 
        \includegraphics[width=0.90\linewidth]{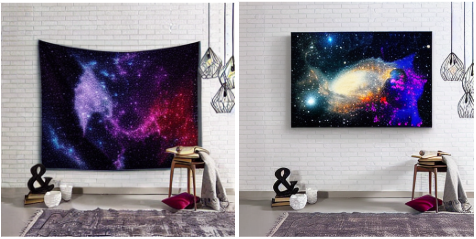}
        \caption*{Background shared between template groups, not suspected for leakage as the source has been found for both. Left prompt: ``Galaxy Wall Tapestry'', right prompt: ``Galaxy Canvas Wall Art Print''.}
        \label{fig:shared_template_not_leakage}
    \end{subfigure}
    \hfill
    \begin{subfigure}{0.48\linewidth}
        \centering
        \vspace{4pt} 
        \includegraphics[width=0.90\linewidth]{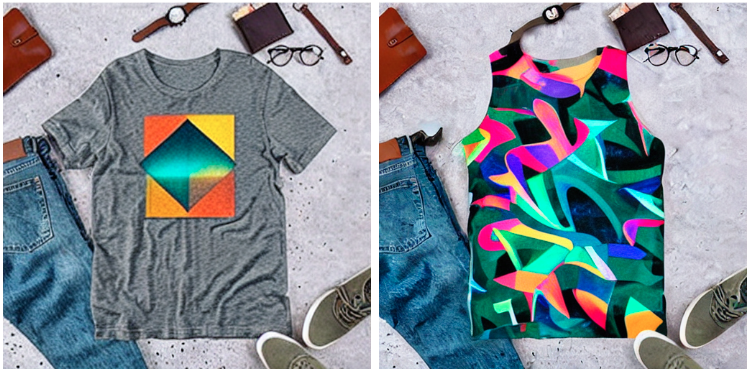}
        \caption*{Suspected leakage from ``T-Shirt'' to ``Tank Top''. Left prompt: ``Abstract Art Essential T-Shirt'', a source has been found. Right prompt: ``Abstract Art Tank Top'', a source has not been found.}
        \label{fig:leakage_tshirt_tank_top}
    \end{subfigure}

    \caption{Example of a template shared between image categories in the training data, and suspected leakage between image categories.}
\end{figure}

Another phenomenon we identified is what we term template leakage: cases where an image template belonging to one template group unexpectedly appears in generations prompted from another template group—or even from a prompt not associated with any template group at all. In these leaked generations, the reused template typically overlaps with the edited region in a visually coherent way. An illustrative example is shown in \cref{fig:leakage_tshirt_tank_top}, where a template strongly associated with the “T-Shirt” template group appeared in a generation prompted for a “Tank Top.” Although the edited region correctly depicts a tank top, the surrounding background matches that of the T-shirt template observed in the source and in other reconstructed images.

However, inspecting e-commerce websites reveals that some templates are genuinely reused across product categories. As a result, not every suspicious case necessarily represents true leakage. For instance, \cref{fig:shared_template_not_leakage} shows a background shared by both “Canvas Wall Art Print” and “Wall Tapestry” because the corresponding image template appears with both categories in the training data.

To differentiate actual leakage from legitimate multi-category reuse, we generated images from our fine-tuned model using prompts that were semantically similar (“Tea Cup”) and semantically dissimilar (“T-Shirt”) to the target category “Coffee Mug.” These additional generations revealed clear evidence of genuine leakage. In particular, with prompts such as “skg Tea Cup,” all resulting images displayed backgrounds that were either interpolations of known coffee-mug templates or matched their color and texture structure. Notably, the generated tea cups consistently appeared in top-view, a perspective absent from the training data and from all coffee-mug generations (which only exhibited the three canonical views). The cups also appeared in shapes not present in the train set. Taken together, these observations confirm that template leakage does occur, producing outputs that combine template structure from one template group with object semantics from another. 

\section{Discussion}
This work highlights that template-level memorization can manifest in ways that evade existing detection methods, particularly through interpolation, perturbations, and cross-category leakage, where copied content is not captured by standard similarity metrics. This limitation stems from the fact that commonly used metrics (e.g., $\ell_2$, SSCD) operate on full-image, pairwise comparisons, where the dominant signal often comes from the editable region, while the copied template occupies only a localized subset; moreover, in interpolation settings, copied elements may originate from multiple sources, further reducing pairwise similarity to any single reference. While avoiding access to the training data better reflects realistic model usage and downstream risk, it also limits the precision with which memorized samples can be identified and attributed. As a result, our analysis relies on manual inspection and domain knowledge, which currently constrains scalability. Developing automated and scalable methods for detecting non-verbatim, template-level memorization remains an important direction for future work, both for understanding memorization mechanisms and for mitigating unintended reuse in deployed generative systems.

\section*{Impact Statement}

This work studies memorization phenomena in text-to-image diffusion models and demonstrates a risk that arises during everyday model usage by naïve users, rather than in adversarial or purely theoretical settings. We show that benign, naturally occurring prompts can lead to the reconstruction of memorized visual content through standard model interfaces, highlighting a gap between existing threat models and real-world usage.

The goal of this research is to promote a clearer understanding of how memorization-related risks manifest in practice, and to support the development of improved evaluation, mitigation, and data curation strategies. By characterizing these behaviors empirically, this work aims to inform model developers, practitioners, and policymakers about potential failure modes that may otherwise go unnoticed.

The methods presented do not provide new access to training data beyond what is already exposed through normal model interaction, and we do not release reconstructed personal data or tools intended for large-scale exploitation. Instead, this work emphasizes responsible disclosure and the importance of addressing memorization risks as text-to-image models continue to be widely deployed.

Overall, we believe that identifying and contextualizing risks that already occur in real-world usage is a necessary step toward safer and more trustworthy deployment of generative models.
\section*{Acknowledgments}

 This project has received funding from the European Research Council (ERC) under the European Union’s Horizon 2020 research and innovation program (grant agreement FoG-101116258).
 Views and opinions expressed are however those of the author(s) only and do not necessarily reflect those of the European Union or the European Research Council. Neither the European Union nor the granting authority can be held responsible for them.
 This work received additional support from the Tel Aviv University Center for AI and Data Science (TAD) and a grant from the Israeli Council of Higher Education.

\FloatBarrier
\bibliographystyle{abbrvnat}
\bibliography{references}

\begin{thebibliography}{41}
\providecommand{\natexlab}[1]{#1}
\providecommand{\url}[1]{\texttt{#1}}
\expandafter\ifx\csname urlstyle\endcsname\relax
  \providecommand{\doi}[1]{doi: #1}\else
  \providecommand{\doi}{doi: \begingroup \urlstyle{rm}\Url}\fi

\bibitem[Aithal et~al.(2024)Aithal, Maini, Lipton, and Kolter]{aithal2024understanding}
S.~K. Aithal, P.~Maini, Z.~C. Lipton, and J.~Z. Kolter.
\newblock Understanding hallucinations in diffusion models through mode interpolation.
\newblock In \emph{Advances in Neural Information Processing Systems (NeurIPS)}, pages 134614--134644, 2024.

\bibitem[Attias et~al.(2024)Attias, Dziugaite, Haghifam, Livni, and Roy]{attias2024information}
I.~Attias, G.~K. Dziugaite, M.~Haghifam, R.~Livni, and D.~M. Roy.
\newblock Information complexity of stochastic convex optimization: Applications to generalization and memorization, 2024.

\bibitem[{Black Forest Labs}()]{FluxSchnell}
{Black Forest Labs}.
\newblock {FLUX.1-schnell} checkpoint.
\newblock URL \url{https://huggingface.co/black-forest-labs/FLUX.1-schnell}.

\bibitem[Carlini et~al.(2021)Carlini, Tram{\`e}r, Wallace, Jagielski, Herbert-Voss, Lee, Roberts, Brown, Song, Erlingsson, Oprea, and Raffel]{carlini2021extracting}
N.~Carlini, F.~Tram{\`e}r, E.~Wallace, M.~Jagielski, A.~Herbert-Voss, K.~Lee, A.~Roberts, T.~Brown, D.~Song, {\'U}.~Erlingsson, A.~Oprea, and C.~Raffel.
\newblock Extracting training data from large language models.
\newblock In \emph{Proceedings of the 30th USENIX Security Symposium (USENIX Security)}, pages 2633--2650, 2021.

\bibitem[Carlini et~al.(2023)Carlini, Hayes, Nasr, Jagielski, Sehwag, Tram{\`e}r, Balle, Ippolito, and Wallace]{carlini2023}
N.~Carlini, J.~Hayes, M.~Nasr, M.~Jagielski, V.~Sehwag, F.~Tram{\`e}r, B.~Balle, D.~Ippolito, and E.~Wallace.
\newblock Extracting training data from diffusion models.
\newblock In \emph{Proceedings of the 32nd USENIX Security Symposium (USENIX Security)}, pages 5253--5270, 2023.

\bibitem[Chen et~al.(2025)Chen, Liu, Shah, and Xu]{chen2025exploringlocalmemorizationdiffusion}
C.~Chen, D.~Liu, M.~Shah, and C.~Xu.
\newblock Exploring local memorization in diffusion models via bright ending attention.
\newblock In \emph{Proceedings of the 13th International Conference on Learning Representations (ICLR)}, 2025.

\bibitem[Chiba-Okabe and Su(2025)]{chiba2025tackling}
H.~Chiba-Okabe and W.~J. Su.
\newblock Tackling copyright issues in {AI} image generation through originality estimation and genericization.
\newblock \emph{Scientific Reports}, 15\penalty0 (1):\penalty0 10621, 2025.
\newblock \doi{10.1038/s41598-025-91790-9}.

\bibitem[{CompVis}()]{SD14HuggingFace}
{CompVis}.
\newblock Stable diffusion version 1.4 checkpoint.
\newblock URL \url{https://huggingface.co/CompVis/stable-diffusion-v1-4}.

\bibitem[{DeepFloyd}()]{DF1HuggingFace}
{DeepFloyd}.
\newblock {DeepFloyd IF-I-XL-v1.0} checkpoint.
\newblock URL \url{https://huggingface.co/DeepFloyd/IF-I-XL-v1.0}.

\bibitem[Elkin-Koren et~al.(2024)Elkin-Koren, Hacohen, Livni, and Moran]{elkin2023can}
N.~Elkin-Koren, U.~Hacohen, R.~Livni, and S.~Moran.
\newblock Can copyright be reduced to privacy?
\newblock In \emph{Proceedings of the 5th Symposium on Foundations of Responsible Computing (FORC)}, 2024.

\bibitem[Feldman(2020)]{feldman2020does}
V.~Feldman.
\newblock Does learning require memorization? a short tale about a long tail.
\newblock In \emph{Proceedings of the 52nd Annual ACM SIGACT Symposium on Theory of Computing (STOC)}, pages 954--959, 2020.

\bibitem[Fredrikson et~al.(2015)Fredrikson, Jha, and Ristenpart]{fredrikson2015model}
M.~Fredrikson, S.~Jha, and T.~Ristenpart.
\newblock Model inversion attacks that exploit confidence information and basic countermeasures.
\newblock In \emph{Proceedings of the 22nd ACM SIGSAC Conference on Computer and Communications Security (CCS)}, pages 1322--1333, 2015.
\newblock \doi{10.1145/2810103.2813677}.

\bibitem[Hacohen et~al.(2024)Hacohen, Haviv, Sarfaty, Friedman, Elkin-Koren, Livni, and Bermano]{hacohen2024not}
U.~Hacohen, A.~Haviv, S.~Sarfaty, B.~Friedman, N.~Elkin-Koren, R.~Livni, and A.~H. Bermano.
\newblock Not all similarities are created equal: Leveraging data-driven biases to inform {GenAI} copyright disputes, 2024.

\bibitem[Haim et~al.(2022)Haim, Vardi, Yehudai, Shamir, and Irani]{haim2022reconstructing}
N.~Haim, G.~Vardi, G.~Yehudai, O.~Shamir, and M.~Irani.
\newblock Reconstructing training data from trained neural networks.
\newblock In \emph{Advances in Neural Information Processing Systems (NeurIPS)}, pages 22911--22924, 2022.

\bibitem[Haviv et~al.(2024)Haviv, Sarfaty, Hacohen, Elkin-Koren, Livni, and Bermano]{haviv2024not}
A.~Haviv, S.~Sarfaty, U.~Hacohen, N.~Elkin-Koren, R.~Livni, and A.~H. Bermano.
\newblock Not every image is worth a thousand words: Quantifying originality in stable diffusion, 2024.

\bibitem[Hintersdorf et~al.(2024)Hintersdorf, Struppek, Kersting, Dziedzic, and Boenisch]{hintersdorf24nemo}
D.~Hintersdorf, L.~Struppek, K.~Kersting, A.~Dziedzic, and F.~Boenisch.
\newblock Finding {NeMo}: Localizing neurons responsible for memorization in diffusion models.
\newblock In \emph{Advances in Neural Information Processing Systems (NeurIPS)}, 2024.

\bibitem[Kandpal et~al.(2022)Kandpal, Wallace, and Raffel]{kandpal2022deduplicating}
N.~Kandpal, E.~Wallace, and C.~Raffel.
\newblock Deduplicating training data mitigates privacy risks in language models.
\newblock In \emph{Proceedings of the 39th International Conference on Machine Learning (ICML)}, pages 10697--10707, 2022.

\bibitem[Lee et~al.(2022)Lee, Ippolito, Nystrom, Zhang, Eck, Callison-Burch, and Carlini]{lee2021deduplicating}
K.~Lee, D.~Ippolito, A.~Nystrom, C.~Zhang, D.~Eck, C.~Callison-Burch, and N.~Carlini.
\newblock Deduplicating training data makes language models better.
\newblock In \emph{Proceedings of the 60th Annual Meeting of the Association for Computational Linguistics (ACL)}, pages 8424--8445, 2022.

\bibitem[Livni(2023)]{livni2023information}
R.~Livni.
\newblock Information theoretic lower bounds for information theoretic upper bounds.
\newblock In \emph{Advances in Neural Information Processing Systems (NeurIPS)}, pages 37716--37727, 2023.

\bibitem[Livni et~al.(2024)Livni, Moran, Nissim, and Pabbaraju]{livni2024credit}
R.~Livni, S.~Moran, K.~Nissim, and C.~Pabbaraju.
\newblock Credit attribution and stable compression.
\newblock In \emph{Advances in Neural Information Processing Systems (NeurIPS)}, pages 2663--2685, 2024.

\bibitem[Ma et~al.(2024)Ma, Cao, Xiao, Li, Zhang, Ye, and Zhao]{ma2024jailbreaking}
J.~Ma, A.~Cao, Z.~Xiao, Y.~Li, J.~Zhang, C.~Ye, and J.~Zhao.
\newblock Jailbreaking prompt attack: A controllable adversarial attack against diffusion models, 2024.

\bibitem[{mattmdjaga}(2024)]{segformerClothesCkpt}
{mattmdjaga}.
\newblock Segformer-b2 clothes fine-tuned checkpoint, 2024.
\newblock URL \url{https://huggingface.co/mattmdjaga/segformer_b2_clothes}.

\bibitem[{Meta AI}(2022)]{maskFormer}
{Meta AI}.
\newblock Maskformer model trained on {ADE20K}, 2022.
\newblock URL \url{https://huggingface.co/facebook/maskformer-swin-tiny-ade}.

\bibitem[{Midjourney}()]{Midjourney}
{Midjourney}.
\newblock Midjourney.
\newblock URL \url{https://www.midjourney.com/}.

\bibitem[{OpenAI}(2022)]{openai2022}
{OpenAI}.
\newblock {DALL-E} 2 pre-training mitigations, 2022.
\newblock URL \url{https://openai.com/index/dall-e-2-pre-training-mitigations/}.

\bibitem[{poloclub}()]{SDprompts}
{poloclub}.
\newblock {DiffusionDB} dataset.
\newblock URL \url{https://huggingface.co/datasets/poloclub/diffusiondb}.

\bibitem[Scheffler et~al.(2022)Scheffler, Tromer, and Varia]{scheffler2022formalizing}
S.~Scheffler, E.~Tromer, and M.~Varia.
\newblock Formalizing human ingenuity: A quantitative framework for copyright law's substantial similarity.
\newblock In \emph{Proceedings of the 2022 Symposium on Computer Science and Law (CSLAW)}, pages 37--49, 2022.
\newblock \doi{10.1145/3511265.3550444}.

\bibitem[Schuhmann et~al.(2022)Schuhmann, Beaumont, Vencu, Gordon, Wightman, Cherti, Coombes, Katta, Mullis, Wortsman, Schramowski, Kundurthy, Crowson, Schmidt, Kaczmarczyk, and Jitsev]{schuhmann2022laion5b}
C.~Schuhmann, R.~Beaumont, R.~Vencu, C.~Gordon, R.~Wightman, M.~Cherti, T.~Coombes, A.~Katta, C.~Mullis, M.~Wortsman, P.~Schramowski, S.~Kundurthy, K.~Crowson, L.~Schmidt, R.~Kaczmarczyk, and J.~Jitsev.
\newblock {LAION-5B}: An open large-scale dataset for training next generation image-text models.
\newblock In \emph{Advances in Neural Information Processing Systems (NeurIPS)}, 2022.

\bibitem[Somepalli et~al.(2023{\natexlab{a}})Somepalli, Singla, Goldblum, Geiping, and Goldstein]{Somepalli2023}
G.~Somepalli, V.~Singla, M.~Goldblum, J.~Geiping, and T.~Goldstein.
\newblock Diffusion art or digital forgery? investigating data replication in diffusion models.
\newblock In \emph{Proceedings of the IEEE/CVF Conference on Computer Vision and Pattern Recognition (CVPR)}, pages 6048--6058, 2023{\natexlab{a}}.

\bibitem[Somepalli et~al.(2023{\natexlab{b}})Somepalli, Singla, Goldblum, Geiping, and Goldstein]{somepalli2023b}
G.~Somepalli, V.~Singla, M.~Goldblum, J.~Geiping, and T.~Goldstein.
\newblock Understanding and mitigating copying in diffusion models.
\newblock In \emph{Advances in Neural Information Processing Systems (NeurIPS)}, 2023{\natexlab{b}}.

\bibitem[{Stability AI}()]{SD35HuggingFace}
{Stability AI}.
\newblock Stable diffusion 3.5 medium checkpoint.
\newblock URL \url{https://huggingface.co/stabilityai/stable-diffusion-3.5-medium}.

\bibitem[Stenbit et~al.(2022)Stenbit, Chollet, and Wood]{kerastutorial}
I.~Stenbit, F.~Chollet, and L.~Wood.
\newblock A walk through latent space with stable diffusion, 2022.
\newblock URL \url{https://keras.io/examples/generative/random_walks_with_stable_diffusion/}.

\bibitem[Tram{\`e}r et~al.(2024)Tram{\`e}r, Kamath, and Carlini]{tramer2024position}
F.~Tram{\`e}r, G.~Kamath, and N.~Carlini.
\newblock Position: Considerations for differentially private learning with large-scale public pretraining.
\newblock In \emph{Proceedings of the 41st International Conference on Machine Learning (ICML)}, pages 48453--48467, 2024.

\bibitem[{vivym}()]{MJprompts}
{vivym}.
\newblock Midjourney prompts dataset.
\newblock URL \url{https://huggingface.co/datasets/vivym/midjourney-prompts}.

\bibitem[Voitovych et~al.(2025)Voitovych, Haghifam, Attias, Dziugaite, Livni, and Roy]{voitovych2025dichotomy}
S.~Voitovych, M.~Haghifam, I.~Attias, G.~K. Dziugaite, R.~Livni, and D.~M. Roy.
\newblock On the dichotomy between privacy and traceability in {$\ell_p$} stochastic convex optimization, 2025.

\bibitem[Webster(2023)]{webster2023}
R.~Webster.
\newblock A reproducible extraction of training images from diffusion models, 2023.

\bibitem[Webster et~al.(2023)Webster, Rabin, Simon, and Jurie]{webster2023deduplicationlaion2b}
R.~Webster, J.~Rabin, L.~Simon, and F.~Jurie.
\newblock On the de-duplication of {LAION-2B}, 2023.

\bibitem[Wolf(2018)]{jackstrattinsmith}
C.~Wolf.
\newblock The accidental model of the alt-right.
\newblock \href{https://www.gq.com/story/accidental-model-redbubble-alt-right-t-shirts}{GQ article}, 2018.

\bibitem[Yang et~al.(2024)Yang, Hui, Yuan, Gong, and Cao]{yang2024sneakyprompt}
Y.~Yang, B.~Hui, H.~Yuan, N.~Gong, and Y.~Cao.
\newblock Sneakyprompt: Jailbreaking text-to-image generative models.
\newblock In \emph{Proceedings of the 45th IEEE Symposium on Security and Privacy (IEEE S\&P)}, pages 897--912, 2024.
\newblock \doi{10.1109/SP54263.2024.00118}.

\bibitem[Yin et~al.(2020)Yin, Molchanov, Alvarez, Li, Mallya, Hoiem, Jha, and Kautz]{yin2020dreaming}
H.~Yin, P.~Molchanov, J.~M. Alvarez, Z.~Li, A.~Mallya, D.~Hoiem, N.~K. Jha, and J.~Kautz.
\newblock Dreaming to distill: Data-free knowledge transfer via deepinversion.
\newblock In \emph{Proceedings of the IEEE/CVF Conference on Computer Vision and Pattern Recognition (CVPR)}, pages 8715--8724, 2020.

\bibitem[Zhang et~al.(2024)Zhang, Jia, Chen, Chen, Zhang, Liu, Ding, and Liu]{zhang2024generate}
Y.~Zhang, J.~Jia, X.~Chen, A.~Chen, Y.~Zhang, J.~Liu, K.~Ding, and S.~Liu.
\newblock To generate or not? safety-driven unlearned diffusion models are still easy to generate unsafe images... for now.
\newblock In \emph{Proceedings of the European Conference on Computer Vision (ECCV)}, pages 385--403, 2024.
\newblock \doi{10.1007/978-3-031-72970-6_22}.

\end{thebibliography}

\clearpage
\appendix
\appendix
\section{Identified Template Groups}\label{apx:template_groups}
Here we list the template groups that we identified through our method described on \ref{sec:our_attack}.

For convenience, the collocations are grouped by product category and the categories are ordered alphabetically. 

The full list of collocations we identified as leading to the reconstruction of partially memorized visual content, is available on our git repository as a .txt file for convenient use. 

\includepdf[pages=-]{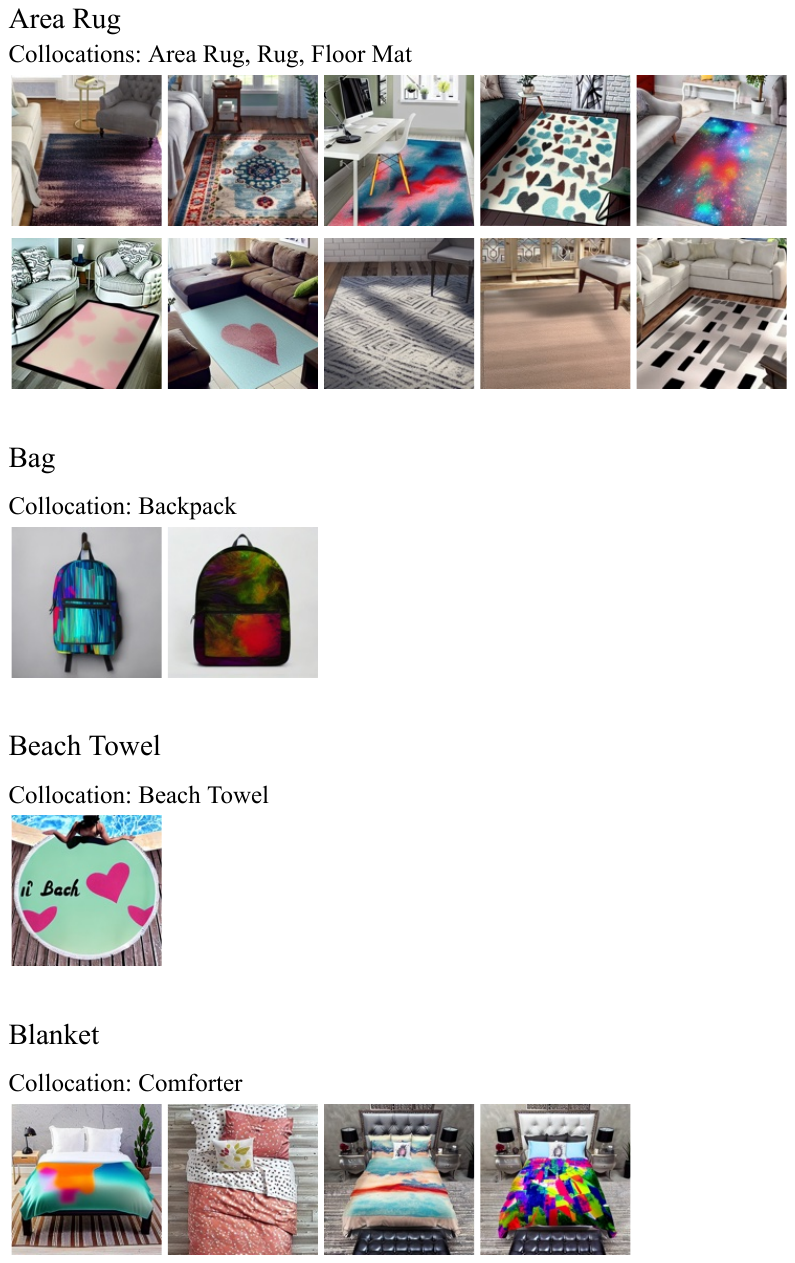}

\end{document}